\documentclass[lettersize,journal]{IEEEtran}
\usepackage{amsmath,amsfonts,amssymb}
\usepackage{algorithmic}
\usepackage{algorithm}
\usepackage{array}
\usepackage[caption=false,font=normalsize,labelfont=sf,textfont=sf]{subfig}
\usepackage{textcomp}
\usepackage{stfloats}
\usepackage{url}
\usepackage{verbatim}
\usepackage{graphicx}
\usepackage{cite}
\usepackage{tabularx}
\usepackage{booktabs}  % 你用了 \toprule, \midrule, \bottomrule
\usepackage{multirow}  % 你用了 \multirow

\newtheorem{theorem}{Theorem} %[section]      % 定理
\newtheorem{corollary}{Corollary}%[section]  % 推论
\newtheorem{lemma}{Lemma}%[section]       % 引理
\renewcommand{\arraystretch}{1.0} % 全局调整行间距，1.0 为默认值

\begin{document}

\title{Physics-Guided Null-Space Diffusion with Sparse Masking for Corrective Sparse-View CT  Reconstruction}

\author{Zekun Zhou$^{2}$, Yanru Gong$^{1}$,  Liu Shi$^{1}$, Qiegen Liu$^{1}$,~\IEEEmembership{Senior Member,~IEEE}
        % <-this % stops a space 
\thanks{This work was supported by National Natural Science Foundation of China (621220033, 62201193). }
\thanks{This work was supported by the Nanchang University Youth Talent Training Innovation Fund Project (Grant: 202506030012).}
%\thanks{This study was funded by the Open Project of the Key Laboratory of Advanced Medical Imaging and Intelligent Computing of Guizhou Province (No. AMIIC250102)) }
\thanks{This work was supported by data from YOFO Technology Co.,Ltd.(Hefei, China).}
\thanks{This work was supported by data from the Institute of Jinan Laboratory of Applied Nuclear Science.}
\thanks{$^{1}$School of Information Engineering, Nanchang University, Nanchang, China.}%
\thanks{$^{2}$School of Mathematics and Computer Sciences, Nanchang University, Nanchang, China. Z. Zhou is the first author (ZekunZhou@email.ncu.edu.cn).}%
\thanks{Corresponding authors: L. Shi (shiliu@ncu.edu.cn) and Q. Liu (liuqiegen@ncu.edu.cn).}%
%\thanks{Z. Zhou is with School of Mathematics and Computer Sciences, Nanchang University, Nanchang, China.}
%\thanks{ L. Shi, Q. Liu and Yanru Gong are with School of Information Engineering, Nanchang University, Nanchang, China. ({shiliu}@ncu.edu.cn, {liuqiegen}@ncu.edu.cn) (Z. Zhou is the first authors.) (Corresponding authors: Liu Shi, Q. Liu)}
% <-this % stops a space
}
%\thanks{Manuscript received April 19, 2021; revised August 16, 2021.}}

\markboth{IEEE TRANSACTIONS ON PATTERN ANALYSIS AND MACHINE INTELLIGENCE, VOL. XX, NO. XX }%
{Shell \MakeLowercase{\textit{et al.}}: A Sample Article Using IEEEtran.cls for IEEE Journals}

\IEEEpubid{~\copyright~2025 The Authors. This work is licensed under a Creative Commons Attribution-NonCommercial-NoDerivatives 4.0 License.}

\maketitle

\begin{abstract}
Diffusion models demonstrate significant advantages in high-fidelity image generation, but their generated results may still deviate from the true data distribution. In medical image reconstruction, such deviations can compromise projection data consistency, causing severe artifacts and substantially undermining reconstruction accuracy and controllability.  To address the above issues,  we propose a physics-guided sparse condition temporal reweighted integrated distribution corrective null-space diffusion model (STRIDE) for sparse-view CT reconstruction.  Specifically,  we design a joint training mechanism guided by sparse conditional probabilities to facilitate the model effective learning of missing projection view completion and global information modeling. Based on systematic theoretical analysis, we propose a temporally varying sparse condition reweighting guidance strategy to dynamically adjusts weights during the progressive denoising process from pure noise to the real image, enabling the model to progressively perceive sparse-view information. The linear regression is employed to correct distributional shifts between known and generated data, mitigating inconsistencies arising during the guidance process. Furthermore, we construct a global dual-network parallel architecture in the wavelet domain to jointly correct and optimize multiple sub-frequency components, effectively mitigating the amplification of frequency deviations during backprojection and enabling high-quality image reconstruction. Experimental results on both public and real datasets demonstrate that the proposed method achieves the best improvement of 2.58 dB in PSNR, increase of 2.37\% in SSIM, and  reduction of 0.236 in MSE compared to the best-performing baseline methods. The reconstructed images exhibit excellent generalization and robustness in terms of structural consistency, detail restoration, and artifact suppression.
\end{abstract}


\section{Introduction}\label{sec:introduction}

\IEEEPARstart{X}{-ray}  computed tomography (CT) as an imaging technique that reconstructs cross-sectional information of an object by acquiring X-ray projections from multiple views\cite{cormack1963representation, hounsfield1973computerized}. Due to the ionizing radiation risks associated with X-rays, minimizing radiation dose has become a central concern in clinical CT imaging. Sparse-view acquisition \cite{brenner2007computed} has emerged as an effective dose-reduction strategy and has been applied in various tasks, such as lung cancer screening \cite{gao2014low}, myocardial perfusion imaging \cite{enjilela2018ultra}. However, the severe loss of projection data continues to pose significant challenges for achieving high-quality image reconstruction and accurate information recovery \cite{whiting2006properties, li2023sparse}.

To mitigate the ill-posed inverse problem caused by missing projection angles, various reconstruction methods have been developed. Traditional approaches such as FBP \cite{pan2009commercial} enable fast reconstruction but suffer from severe streak artifacts, while iterative methods (ART \cite{gordon1970algebraic}, PWLS \cite{fessler1994penalized}, EM \cite{dempster1977maximum}) improve stability but rely on strong prior information. Compressed sensing \cite{donoho2006compressed} and TV minimization \cite{yu2005total,sidky2008image} methods exploit sparsity or regularization to enhance quality but often struggle with complex structures or introduce edge blurring. CNN \cite{han2018framing} and GAN \cite{li2019sinogram} deep learning approaches can learn data-driven priors, yet their generalization is limited and global artifact structures are difficult to capture. Recently, diffusion models have shown significant advantages in sparse-view CT reconstruction \cite{ho2020denoising, song2020score, yang2025ct}. These approaches can be broadly categorized into image-domain prior modeling \cite{wu2024multi, wu2024linear}, which integrates data consistency and   multi-level diffusion to enhance stability, and projection domain \cite{xia2022patch, guan2023generative} or dual-domain \cite{yang2024dual} collaborative reconstruction, which leverages sinogram refinement, patch-based denoising, or multi-channel generative modeling to infer missing data and suppress artifacts. 

\IEEEpubidadjcol

Although guided diffusion models demonstrate strong controllability, their guidance signals are usually derived from models or conditions trained under specific distributions, which inherently differ from the target domain and thus inevitably induce distribution shifts \cite{ho2022classifier}. Several strategies have been proposed to mitigate this issue, such as semantic-aware guidance \cite{shen2024rethinking}, feedback guidance \cite{felix2025feedback}, and self-degraded replica-based guidance \cite{karras2024guiding}, which have proven effective in alleviating distribution shifts for natural images and cross-modal generation. However, these methods do not take into account the unique characteristics of CT projection-domain reconstruction. In this setting, the intrinsic limitation of diffusion models in high-frequency information modeling is further amplified, making it difficult for local detector measurements to achieve globally consistent mappings during backprojection, thereby causing severe artifacts and detail loss in the reconstructed images. Against this backdrop, sparse-view scanning is particularly challenging, as the severe lack of projection angles exacerbates both distribution shifts and the limitations in high-frequency recovery, which can severely degrade reconstruction quality. Nevertheless, by reasonably incorporating sparse-view  physical priors into the diffusion process, it is possible to effectively alleviate the adverse effects of distribution shifts and improve reconstruction performance.

To address these limitations, we propose a physics-guided \textbf{S}parse condition \textbf{T}emporal \textbf{R}eweighted  \textbf{I}ntegrated \textbf{D}istribution correctiv\textbf{E} null-space diffusion model (STRIDE). The model uses joint training with sparse conditional probability guidance to complete missing views and capture global data distribution.  During reconstruction, based on a theoretical analysis of sparse data guidance mechanisms, a time-varying sparse conditional reweighting strategy is systematically designed. By dynamically adjusting the guidance weights, this strategy enables the model to progressively perceive and effectively utilize sparse-view information across different denoising stages. To mitigate potential distribution shifts between generated and known projection data, a linear regression method is applied to calibrate the distributions, ensuring data consistency and authenticity.  Additionally, the model employs a dual-network parallel architecture to optimize and correct multiple sub-frequency components of the image. The proposed method offers a novel and efficient solution for SVCT reconstruction.

The main contributions of this paper can be summarized as follows:

\indent $\bullet$ \emph{\textbf{Joint Prior Learning with Stochastic Masking.}}  We propose a joint training mechanism based on sparse conditional probability guidance, which effectively enables the model to simultaneously complete missing projection views and model the global data distribution, thereby enhancing the accuracy and consistency of SVCT reconstruction.

\indent $\bullet$  \emph{\textbf{Temporal Weighting Control Sampling.}} We developed a temporal guidance weight modulation framework based on sparse-view masks. By integrating theoretical analyses of error propagation and the influence of guidance terms during the sampling process, we systematically derived an optimal dynamic weighting strategy that balances generative diversity and structural fidelity. Based on this, a time-aware guidance function was designed. Experimental results further demonstrate that this theory-driven approach significantly enhances reconstruction quality and structural consistency under sparse-view conditions, validating the effectiveness and practicality of the proposed mechanism.

\indent $\bullet$  \emph{\textbf{Dual-Channel Wavelet-Based Consistency Correction.}} We propose a dual-channel network-based stationary wavelet correction method. Diffusion models often struggle to effectively capture high-frequency information, and such high-frequency inconsistencies are typically amplified during backprojection. To address this, our method performs correction across the global frequency space to explicitly compensate for the diffusion models’ shortcomings in high-frequency modeling, thereby suppressing artifacts and enhancing detail consistency.

In Section II, we briefly review related work. Section III details the theoretical framework and elaborates on our proposed approach. Experimental comparison results are shown in Section IV. Finally, Section V provides discussion and concluding remarks on the methods introduced.

\section{Preliminary}

\subsection{Sparse-view CT Image Reconstruction}

Let $y=[y_{1},y_{2}, \cdots, y_{m} ]^{T}$ denote a projection data vector measured by the detectors. $y_{i}$ is the projection value of the $i$-th detector,  $m$ is the total number of the detectors.  Image reconstruction problem in CT can be mathematically formulated as an inverse problem of solving the following linear equation:
%\vspace{-0.5 em}
\begin{equation}
y= Ax+  \eta,   \label{eq1}
\end{equation}
where $x$ denote a linear attenuation coefficients distribution of the object  to be reconstructed. $A=(a_{ij})\in \mathbb{R}^{m \times n}$ represents the CT system matrix, where each $a_{ij}$ depends on the projection angle, and $\eta \in \mathbb{R}^{m}$ denotes the system noise. The goal is to reconstruct the unknown CT image $x$ from the observed measurement $y$.

SVCT reconstruction problem is a classical inverse problem [23]. Under conditions of high sampling rate and high signal-to-noise ratio, the image $x$ can typically be reconstructed with high quality. However, increasing the number of projection views significantly raises the radiation dose, thereby increasing the potential risk of cancer. Consequently, accurately reconstructing the unknown image from limited measurements becomes a highly challenging task.  The transformation from the full projection data  $y$ to sparse-view data $y_{s}$ can be modeled as a linear transformation.  Specifically, this process can be represented by a linear operator $P(\cdot)$ acting on $y$, which performs row-sparse sampling as a linear mapping function $f:\mathbb{R}^{m\times n} \rightarrow \mathbb{R}^{m\times n}$, expressed as follows:
\begin{equation}
y_{s}=P(y).   \label{eq2}
\end{equation}

Resulting from the intrinsic incompleteness of measured data, the problem becomes ill-posed, and the linear system may have infinitely many solutions, resulting in reconstructed images with artifacts, blurring, and noise. To address this issue, prior information is typically incorporated into the objective function, which is minimized with an L2 norm regularization term, expressed as follows:
\begin{equation}
x^{\ast}=\underset{x}{\arg\min} \ \frac{\lambda}{2} \lVert y-Ax  \rVert ^{2} _{2} + \mu R(x),  \label{eq3}
\end{equation}
where the first term represents data fidelity, ensuring consistency between the undersampled data and the actual measurements; the regularization term $R(x)$ introduces prior constraints and is balanced by the weighting factor $\mu$.

\subsection{Diffusion-Based Generative Models}

Diffusion models as generative models with strong expressiveness and high flexibility \cite{ho2020denoising, sohl2015deep, song2019generative }, have attracted increasing attention in recent years.  Denoising Diffusion Probabilistic Model (DDPM) \cite{ho2020denoising} is one of the earliest diffusion models, which models the data generation process as a Markov chain, using a denoising network at each step to predict the original signal from the noisy input. The forward process is typically modeled as a fixed sequence of Gaussian perturbations, expressed as:
\begin{equation}
q(x_{t}|x_{t-1})=\mathcal{N}(x_{t};\sqrt{1-\beta_{t}}x_{t-1},\beta_{t}\mathbf{I}),  \label{eq4}
\end{equation}
where  $\beta_{t}$ denotes the variance of the added noise. The reverse process is also modeled as a Gaussian distribution, with its parameters predicted by a neural network $\epsilon_{\theta}$.
\begin{equation}
p_{\theta}(x_{t-1}|x_{t})=\mathcal{N}(x_{t-1};\mu_{\theta}(x_{t},t),\Sigma_{\theta}(x_{t},t)).  \label{eq5}
\end{equation}
However, the reverse process in DDPM typically requires hundreds to thousands of sampling steps, resulting in slow generation.  To improve sampling efficiency, DDIM \cite{song2020denoising} proposed a non-Markovian process that reduces the number of steps via deterministic sampling while maintaining generation quality. Building on this, IDDPM \cite{nichol2021improved} further optimized the noise schedule and sampling strategy, enhancing both efficiency and reconstruction performance. To improve controllability, Ho et al. \cite{dhariwal2021diffusion} introduced classifier guidance by leveraging the gradient of a pretrained classifier  $p(y|x)$ to guide the generation process, though this increases training and computational cost. To address this, Classifier-Free Guidance  \cite{ho2022classifier} proposed a classifier-independent strategy that combines conditional and unconditional inputs during training, enabling effective guidance at inference time via interpolation:
\begin{equation}
\tilde{\epsilon}_{\theta}=(1+\omega)\epsilon_{\theta}(x_{t}|\mathbf{c})-\omega \epsilon_{\theta}(x_{t}),  \label{eq6}
\end{equation}
where  $\omega$ denotes the guidance weight, and $\mathbf{c}$ represents the conditional input. This strategy effectively reduces model complexity and training cost.

\section{Methods}

\subsection{Motivation}

Diffusion models have gradually become one of the core methods in SVCT reconstruction due to their superior performance in modeling complex data. However, despite their remarkable ability to generate images with rich details and complex structures, the results produced by diffusion models often exhibit statistical discrepancies from the true images \cite{liu2024correcting}. This inconsistency is especially pronounced in high-precision applications such as medical imaging, where it can lead to structural distortions and reduced reliability. Therefore, mitigating the discrepancy between the generated and true distributions  has become a key challenge in improving the quality of SVCT reconstruction.

\begin{figure}[!t]
\centering
\centerline{\includegraphics[width=0.99\columnwidth]{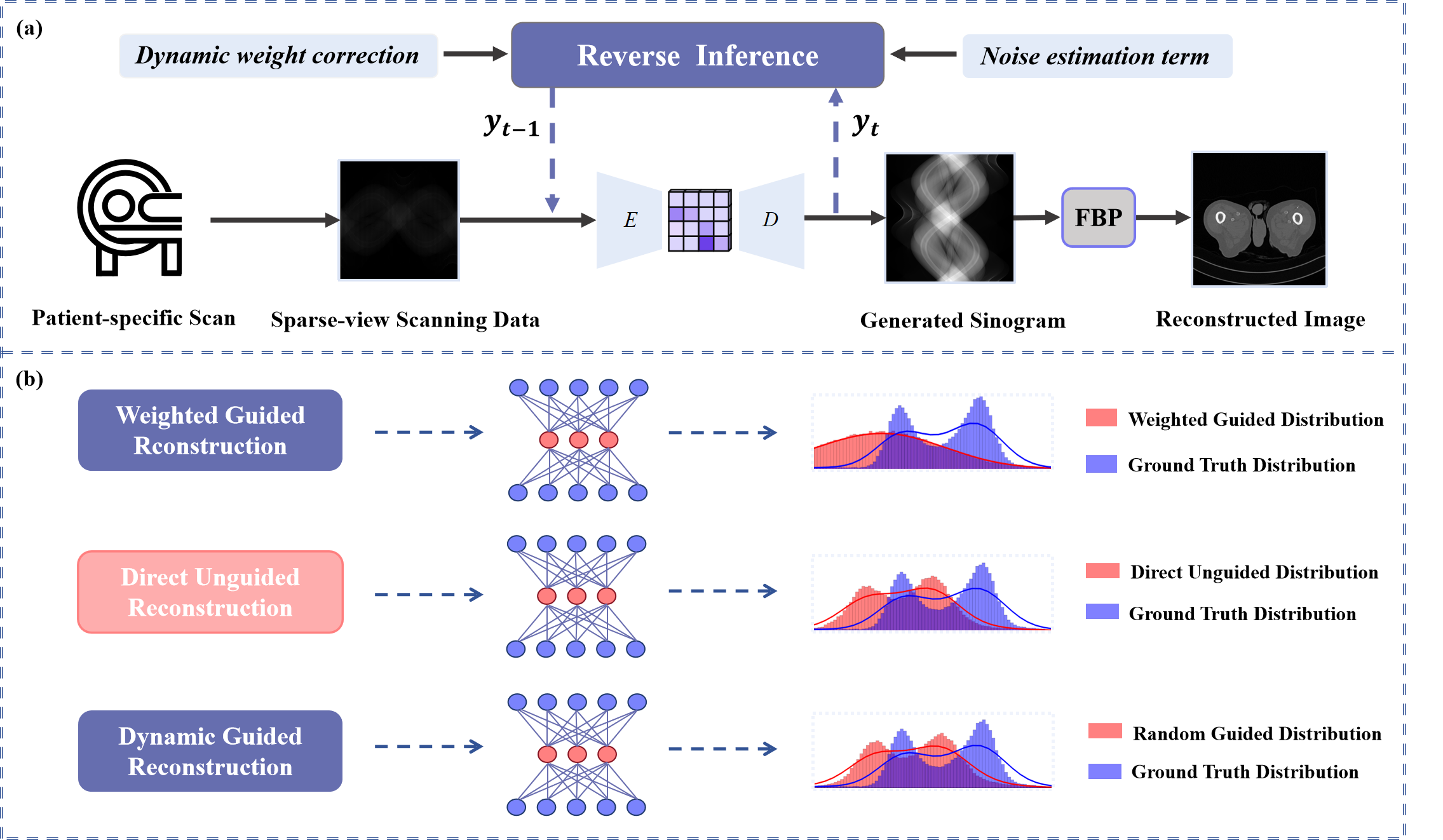}}
\caption{(a) Illustration of the network architecture with guided correction weighting strategy.
(b) Comparison of the distributional differences between generated results under different guided correction weighting strategies and the Ground Truth.}
\label{Motivation}
\end{figure}

Physical priors and structural information are frequently incorporated into diffusion models to improve image fidelity and consistency. However, fixed-strength guidance can perturb latent distribution learning, resulting in semantic shifts and detail degradation \cite{felix2025feedback, gao2025noise}. In the context of SVCT reconstruction, the pronounced sparsity and incompleteness of projection data make fixed-strength guidance particularly inadequate for dynamic diffusion processes.

As shown in Fig. \ref{Motivation}, we address this limitation through a systematic theoretical analysis. Our results (Theorem \ref{thm1}) establish that, at each iteration step, an optimal weighting factor exists that allows the guidance to attain the best reconstruction performance. Motivated by this finding, we introduce a time-step–driven sparse mask dynamic weighting correction strategy, which adaptively modulates guidance strength across iterations to enhance both structural consistency and reconstruction quality. Moreover, we analyze the evolution of the weighting factor with respect to iteration steps (Corollary \ref{cor1}), providing a rigorous theoretical foundation for experimental design and strategy refinement.

\begin{theorem}\label{thm1}
Let $\hat{y}_{0}^{(t)}$ denote the model's estimate of the initial signal at time step $t$, and let $y_{s}$ denote the observed data. The corrected estimate is defined as $\tilde{y}_{0}^{(t)}=\hat{y}_{0}^{(t)}+\lambda_{t}M(\hat{y}_{0}^{(t)}-y_{s})$, where $\lambda_{t} \in [0, 1]$ is a time-dependent dynamic correction weight, $M$ is a sparse mask. There exists an optimal sequence of weights $\lambda^\ast = \{ \lambda_T^\ast, \dots, \lambda_0^\ast \}$ that minimizes the overall reconstruction error under a specified metric.
\end{theorem}

The proof is provided in Appendix.

\begin{corollary}\label{cor1}
Let ${y}_s$ denote the observed data and $y_g$ the ground truth. Define the estimation errors after dynamic correction as $\zeta_{0}^{(t)} = \hat{y}_{0}^{(t)}-y_{g}$ and $\xi_{g}=y_{s}-y_{g}$. Then, the optimal correction weight $\lambda_t^{\ast}$ satisfies the following conditions, under the feasible constraint $\lambda_t \in [0,1]$.  If $|\zeta_{0}^{(t)}|^2 < |\xi_{g}|^2$, then the optimal correction weight $\lambda_t^{\ast}$ typically decreases as the iteration proceeds, reflecting the convergence of the estimation error; otherwise,  a truncation constraint should be imposed to prevent over-correction.
\end{corollary}

Proof is deferred to Appendix.

Although diffusion models have exhibited strong potential in SVCT reconstruction, their capacity to accurately model and recover high-frequency information remains limited \cite{zhao2025frequency}. In projection data, high-frequency components not only encode essential structural details and edge information but also play a pivotal role in preserving the physical consistency of the backprojection process. Any deviations in these components may accumulate, thereby compromising their physical interpretability, exacerbating artifacts, and ultimately degrading reconstruction quality. To overcome these limitations, we introduce a Stationary Wavelet Transform driven correction strategy that employs multi-scale decomposition to independently refine high-frequency and low-frequency components. By enhancing structural detail while ensuring global consistency, the proposed approach effectively mitigates physical inconsistencies in the projection process, leading to substantial improvements in reconstruction quality and stability.

\begin{figure}[!t]
\centerline{\includegraphics[width=1.0\columnwidth]{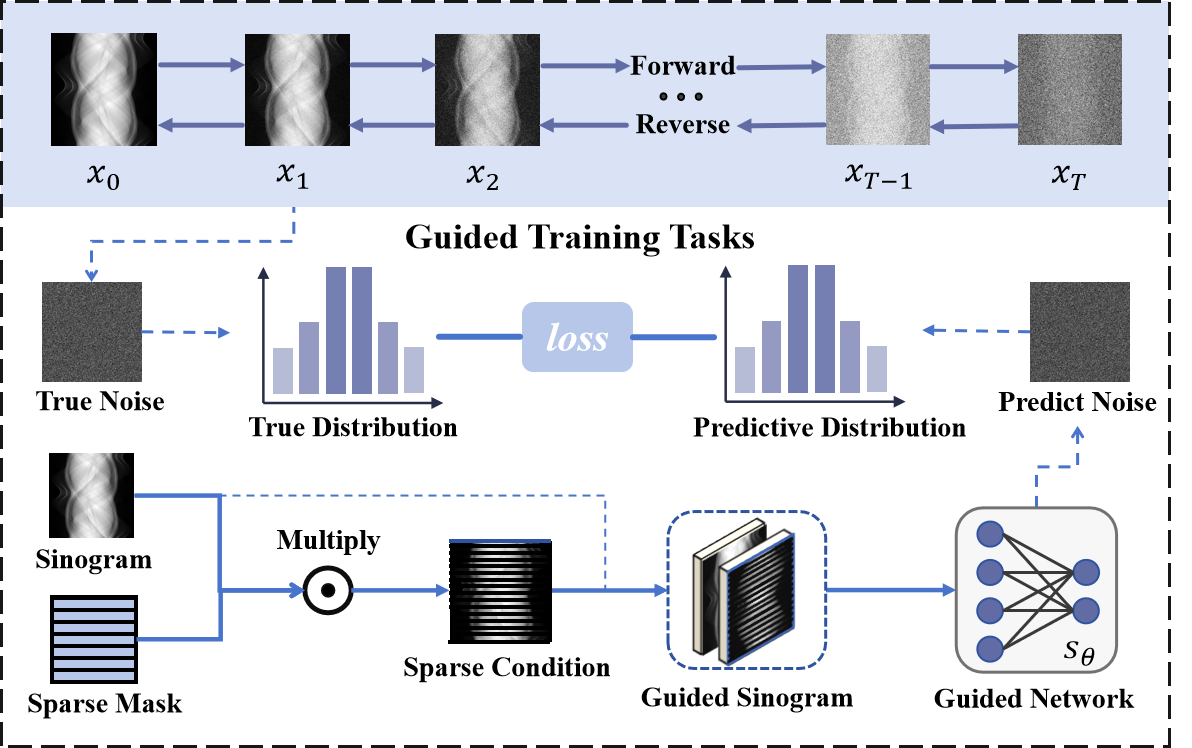}}
\caption{Sparse Mask Embedded Conditional Guided Diffusion. Probabilistically embedded sparse mask into the conditional guided diffusion process, allowing the model to capture global projection information while emphasizing the completion of missing regions.}
\label{GuidedTrainPIC}
\end{figure}

\subsection{Training Process}

\begin{figure}[t!]
\centerline{\includegraphics[width=1.0\columnwidth]{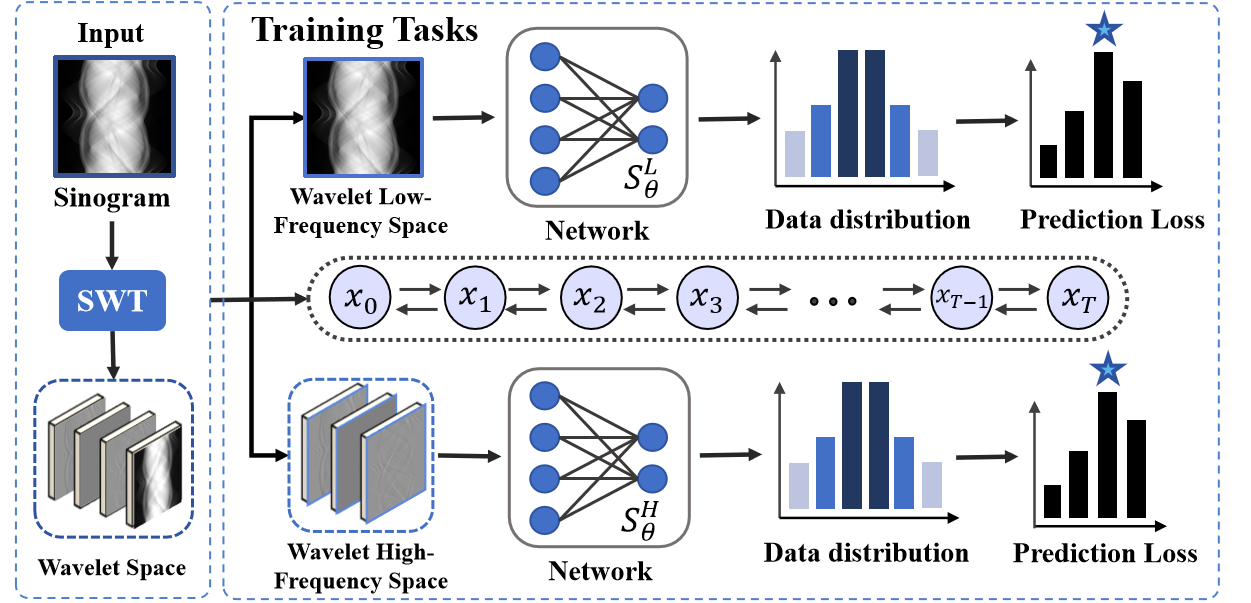}}
\caption{Correction Training Strategy Based on SWT. The sinogram is decomposed by SWT into global low-frequency and detailed high-frequency components, which are trained separately to capture global structures and fine details for distribution correction.}
\label{CorrTrain}
\end{figure}

During training, we propose a diffusion-based correction mechanism for SVCT reconstruction, comprising:  As shown in Fig. \ref{GuidedTrainPIC}, a probabilistic guidance-based conditional diffusion framework, where sparse masks are conditionally applied to improve adaptability to global and local structures; and a dual-branch multi-frequency diffusion strategy, with independent diffusion and score networks for low-frequency and high-frequency components, effectively correcting frequency biases and enhancing detail reconstruction in Fig. \ref{CorrTrain}.

\subsubsection{Sparse Mask Embedded Conditional Guided Diffusion} 

Let the input projection data be denoted by $y_{0} \in \mathbb{R}^{A \times D}$, where $A$ denotes the number of angular views and $D$ represents the number of detector elements.  During the forward diffusion process, a sequence of predefined noise scaling coefficients $\{ \bar{\alpha}_{t} \}_{t=1}^{T}$ is used to progressively inject Gaussian noise into the input projection data at each time step $t\in \{ 1, \cdots, T \}$, resulting in a noisy observation $y_{t}$ . The forward process is formally defined as:
\begin{equation}
y_{t} = \sqrt{\bar{\alpha}_{t}} y_{0} + \sqrt{1 - \bar{\alpha}_t} \epsilon, \quad \epsilon \sim \mathcal{N}(0, \mathbf{I})
\label{eq7} % 标签用于交叉引用，这里标签名为eq7，可根据需要修改
\end{equation}
where $\bar{\alpha}_{t} = \prod_{s=1}^t \alpha_{s} \in (0, 1]$ denotes the cumulative product of noise retention factors up to step $t$, controlling the trade-off between signal preservation and noise injection. The term $\epsilon$ represents isotropic standard Gaussian noise independent of the input signal.

To simulate different sparse-view scanning scenarios, a sparse mask is introduced in the projection domain, defined as:
\begin{equation}
M_{i} = \mathbf{1}_{\{i \not\equiv 0 \pmod{r}\}},
 \label{eq8}
\end{equation}
where $ i $ denotes the angular index in the projection data, and $ r $ is a sampling interval randomly drawn from a predefined view set with a uniform distribution. The mask $ M $ controls the number and distribution of visible views. Define a sampling operator $\mathcal{S}_{M}: \mathbb{R}^{A \times D} \to \mathbb{R}^{A \times D}$ to represent the projection data sampling process conditioned on the mask $M$. Specifically, when applied to the original projection data $y_{0}$, this operator produces the conditional input $\tilde{c}_{M}$:
\begin{equation}
\tilde{c}_{M} = \mathcal{S}_M(y_{0}), \quad \text{where} \quad \mathcal{S}_M(y_{0}) := M \circ y_{0} ,  \label{eq9}
\end{equation}
where  $\circ$ denotes the Hadamard product.

To enhance the model's generalization and flexibility between conditional dependence and unconditional generation, we introduce a probabilistic guidance mechanism during training stage. Specifically, for each noised sample, the use of conditional guidance is determined by a random variable $\gamma \sim \text{Bernoulli}(p)$:
\begin{itemize}
  \item If $\gamma = 1$ (with probability $p$), the conditional guidance based on the angular mask $\tilde{c}_{M}$ is applied;
  \item If $\gamma = 0$ (with probability $1 - p$), the model is trained without any conditional input.
\end{itemize}

The model estimates the original noise in the noisy projection data through a conditional noise prediction network \(\epsilon_{\theta}(\cdot)\). When conditional guidance is introduced, the network learns the conditional noise distribution given a specific guidance condition, which can be expressed as:
\begin{equation}
\epsilon_{\theta}(y_{t},t,\gamma \cdot \tilde{c}_{M}) \approx \epsilon \sim q(\epsilon \mid y_{0}).  \label{eq10}
\end{equation}
The reverse diffusion process is modeled as a conditional Gaussian distribution:
\begin{equation}
p_{\theta}(y_{t-1}\mid y_{t}, \gamma \cdot \tilde{c}_{M}):=  \mathcal{N} \bigl (y_{t-1};\mu_{\theta}(y_{t},t,\gamma \cdot \tilde{c}_{M}), \Sigma_{t} \bigr),  \label{eq11}
\end{equation}
where  $\tilde{c}_M$ denotes the guidance condition generated from the random angular mask, $\gamma \in \{0,1\}$ is a Bernoulli-distributed random variable controlling whether conditional input is used, and $\mu_{\theta}(\cdot)$ is the mean term computed by the noise prediction network, defined as:
\begin{equation}
\mu_{\theta}(y_{t},t,\gamma \cdot \tilde{c}_{M})= \frac{1}{\sqrt{\alpha_{t}}} (y_{t}-\frac{1-\alpha_{t}}{\sqrt{(1-\bar{\alpha_{t}})}} \epsilon_{\theta}(y_{t},t, \gamma \cdot \tilde{c}_{M}) ).  \label{eq12}
\end{equation}
Finally, the model training objective is to minimize the noise prediction error, defined by the loss function:
\begin{equation}
\mathcal{L}_{GM}= \mathbb{E}_{y_{0},\epsilon,t,\tilde{c}_{M},\gamma} \left[ \left\| \epsilon - \epsilon_{\theta}(y_{t},t,\gamma \cdot \tilde{c}_{M}) \right\|_{2}^{2} \right].  \label{eq13}
\end{equation}

\subsubsection{Correction Training Strategy Based on SWT} % 随机角度掩码引导式扩散训练

Applying the Stationary Wavelet Transform (SWT) to the fully-sampled projection data $ y_{0} $ to decompose it into low-frequency and high-frequency components. Specifically, the decomposition is defined as:
\begin{equation}
y_{0}^\omega[n] = \sum_{k} f_{\omega}[k] \cdot y_{0}[n - 2^j k], \quad \omega \in \{L, H\},
\label{eq:swt}
\end{equation}
where $f_{\omega}[k]$ represents both the low-frequency and high-frequency convolutions unified filter,  $\omega=L$ and $\omega=H$  corresponds to the low-frequency branch and  high-frequency branch, respectively. A time-continuous diffusion process is introduced to model the dynamic evolution of the data:
\begin{equation}
dy_t^\omega = \sqrt{\frac{d}{dt} \sigma_\omega^2(t)} \, dW_t^\omega, \quad \omega \in \{L, H\},
\label{eq:SDE}
\end{equation}
where $y_{0}^{\omega} \sim p_{\mathrm{data}}^{\omega}$,  $t \in [0,T]$. $\{W_{t}^{\omega}\}_{t \geq 0}$  independent standard Brownian motions, and the diffusion coefficients $\sigma_{\omega}^2(t)$  strictly increasing and continuously differentiable non-negative functions.

Within the diffusion model framework, the training objective is to learn the gradient of the data distribution $p_{t}(y_{t})$ at any given time $t$, known as the score function. Specifically, the model employs a parameterized network $s_{\theta}(y_{t}, t)$ to approximate this score function, $
s_{\theta}(y_{t}, t) \approx \nabla_{y_{t}} \log p_{t}(y_{t})$. $ \nabla_{y_{t}} \log p_{t}(y_{t}) $ denotes the gradient of the log-probability density of the data at time $t$ with respect to  $ y_{t} $. The model is optimized by minimizing the following loss function:
\begin{equation}
\begin{split}
\mathcal{L}_{\omega}(\theta_\omega) &= 
\mathbb{E}_{t \sim U(0,T)} \, 
\mathbb{E}_{y_t^\omega \sim p_{\text{data}}^\omega} 
\Big[ \lambda_\omega(t) \, \| s_{\theta_\omega}(y_t^\omega, t)  \\
&- \nabla_{y_t^\omega} \log p_t^\omega(y_t^\omega) \|_2^2 \Big], \quad \omega \in \{L, H\},
\end{split}
\label{eq:sde_loss}
\end{equation}
where \(\lambda_{\omega}(t)\)  denote the dynamic weighting functions.

\begin{figure*}[!t]
\centering
\includegraphics[width=\textwidth]{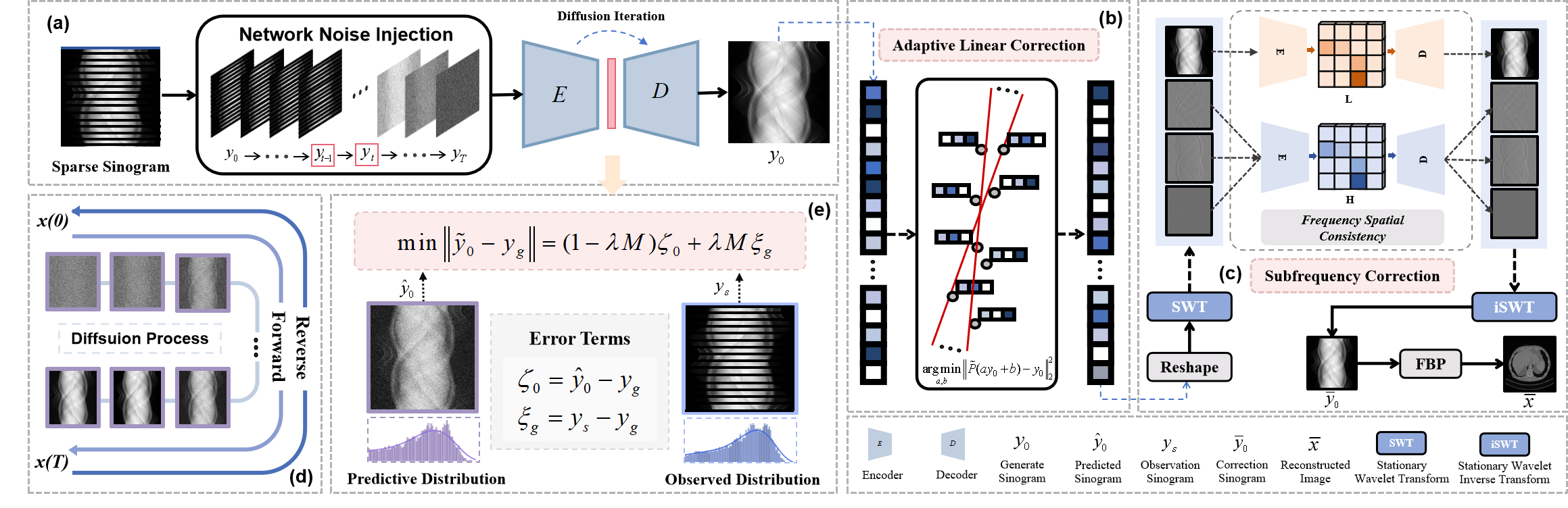} % 使用 \textwidth 占据双栏宽度
\caption{Overview of the proposed sparse-view CT reconstruction framework. (a) A temporally reweighted guided diffusion module provides fast coarse generation of projection data, ensuring global structural consistency under sparse-view conditions. (b) An adaptive linear data-consistency correction is then applied to enforce fidelity with the measured sinogram. (c) The refined projections are further decomposed by stationary wavelet transform (SWT), where low-frequency and high-frequency components are separately corrected by dedicated networks to capture both global trends and fine structural details. (d) Interpretation of diffusion process. (e) Temporal reweighting-guided correction process.}
\label{Test}
\end{figure*}

\subsection{Sinogram Reconstruction and Correction Stage}

During the reconstruction stage, we propose a temporal reweighted distribution correction framework to address data incompleteness and structural degradation caused by sparse sampling.
As shown in Fig. \ref{Test},  This framework applies a theoretically grounded reweighting strategy to the sampling process and adaptively performs linear correction between the original and generated data to improve consistency. Additionally, a dual-branch multi-frequency correction module is introduced to refine different frequency components, enabling cross-domain collaborative compensation and correction. This design ensures global consistency while significantly enhancing reconstruction accuracy and detail fidelity.

\subsubsection{Temporal Reweighted Sampling Correction} %引导式投影数据快速粗生成

We employ the Denoising Diffusion Implicit Model (DDIM) for initial reconstruction, incorporating a skip sampling strategy to reduce the number of iterations and accelerate the sampling process. Specifically, at each discrete time step $ t \in \{ T, T-\Delta, \ldots, 0 \} $, DDIM predicts the denoised initial sample $ y_{0} $ from the current state $ y_{t} $ and performs a backward estimation to obtain the previous state $ y_{t-1}$, progressively approximating the true projection data.

At any given time step $t $, the model estimates the pseudo-original data $\hat{y}_{0}^{(t)}$  using a noise prediction network:
\begin{equation}
\hat{y}_{0}^{(t)} = \frac{1}{\sqrt{\bar{\alpha}_{t}}} \left( y_{t} - \sqrt{1 - \bar{\alpha}_{t}} \, \hat{\epsilon}_{\theta}(y_{t}, t) \right),
\label{eq16}
\end{equation}
where $\hat{\epsilon}_{\theta}(\cdot)$ denotes the network’s estimation of noise,  $\alpha_{t}$ is the pre-defined noise attenuation coefficient. To introduce conditional guidance for correcting the generated projection data, $ M \in \{0,1\}^{n} $ is embedded into the predicted pseudo-original data $\hat{y}_{0}^{(t)}$. The guidance strength $\lambda_{t}$ is dynamically adjusted over time $ t $ and defined as:
\begin{equation}
\lambda_{t} = \min\left(1, \frac{t}{T} \right) \nu, \quad \nu_{\max} = 1,
\label{eq17}
\end{equation}
where $\lambda_{t}$ represents the guidance weight at time step $ t $. The guided and corrected prediction is given by:
\begin{equation}
\tilde{y}_{0}^{(t)} = \hat{y}_{0}^{(t)} - \lambda_{t}  M  \left( y_{s} - \hat{y}_{0}^{(t)} \right),
\label{eq18}
\end{equation}
where $y_{s}$ denotes the observed sparse projection data. Finally, the iterative backward sampling update is formulated as:
\begin{equation}
y_{t-1} = \sqrt{\bar{\alpha}_{t-1}} \tilde{y}_{0} ^{(t)}+ \sqrt{1 - \bar{\alpha}_{t-1}} \cdot \hat{\epsilon}_{\theta}(y_{t}, t) + \sigma_{t}z, 
\label{eq19}
\end{equation}
where $\sigma_{t}$ is the noise control coefficient, $ z \sim \mathcal{N}(0, \mathbf{I}) $ is a standard Gaussian noise.

To mitigate distribution mismatches between generated and measured projection data, we employ an adaptive linear regression strategy that aligns the overall intensity distribution of each generated projection with its corresponding measured data. By independently estimating linear mapping parameters for each projection, this method dynamically adjusts scale and offset to harmonize the generated data distribution with real measurements. This adaptive, per-sample distribution alignment ensures improved numerical consistency and enhances the physical fidelity of reconstructed images. Estimate scaling $a$ and shift $b$ by solving:
\begin{equation}
\{a^{*}, b^{*}\} = \arg\min_{a,b} \| M \tilde{P}(a y_{0} + b) - y_{s} \|_{2}^{2},
\label{eqLinear}
\end{equation}
where $\tilde{P}(\cdot)$ is the projection operator, $y_{0}$ denotes the generated data result under the scenario of distribution shift.

\subsubsection{Correction Mechanism Based on Stationary Wavelets} 

To mitigate frequency shifts introduced by coarse diffusion generation, we decouple the projection data into distinct frequency bands in the wavelet domain, introducing separate modeling processes for high-frequency and low-frequency components. By integrating prior information and consistency constraints, this design facilitates frequency alignment and enhances the recovery of structural details.The overall optimization objective integrates data consistency with frequency-specific regularizations, formally defined as:
\begin{equation}
\bar{y}^{\ast} = \arg\min_{\bar{y}} \left\{ \| P(\bar{y}) - y \|_2^2 + \lambda_{L} R_{L}(\bar{y}^{L}) + \lambda_{H} R_{H}(\bar{y}^{H}) \right\},
\label{eq:optimization}
\end{equation}
where $y$ denotes the observed true projection data, $P(\cdot)$ is the sampling operator, $\bar{y}$ represents the reconstructed complete projection data, $R_{L}(\cdot)$ and $R_{H}(\cdot)$ are the regularization terms for the low-frequency and high-frequency components respectively, and $\lambda_{L}$ and $\lambda_{H}$ are the corresponding regularization weights.

The associated reverse diffusion process can be described by the following stochastic differential equation (SDE):
\begin{equation}
d \bar{y} = \bigl( f(\bar{y}, t) - g^{2}(t) s_{\theta}(\bar{y}, t) \bigr) dt + g(t) d\bar{w},
\label{eq:SDE}
\end{equation}
where $t$ denotes the time parameter decreasing from $T$ to $0$, $\bar{w}$ is a standard Brownian motion, $f(\cdot)$ and $g(\cdot)$ are the drift and diffusion coefficients respectively, and $s_{\theta}(\cdot)$ is the parameterized score function.

For the coarsely generated projection data $\bar{y}$, we first perform a multi-scale decomposition to separate it into low-frequency component \(\bar{y}^{L}\) and high-frequency components $\bar{y}^{H} = \{\bar{y}^{H_{1}}, \bar{y}^{H_{2}}, \bar{y}^{H_{3}}\}$, representing details along different directions. Subsequently, a corrector based on Langevin dynamics is applied to iteratively update these frequency components. Specifically, at iteration $t$, both low-frequency and high-frequency components are updated under the guidance of their respective score networks $ s_{\theta}^{\omega}$,  formulated as:
\begin{equation}
\bar{y}_{t-1}^{\omega} = \bar{y}_{t}^{\omega} + \epsilon_{t} s_{\theta}^{\omega}(\bar{y}_{t}^{\omega}, t) + \sqrt{2 \epsilon_{t}} z_{t}^{\omega}, \quad \omega \in \{L, H\},
\label{eq23}
\end{equation}
where $\epsilon_{t} > 0$ is the step size, and $z_{t}^{\omega} \sim \mathcal{N}(0, I)$  independent Gaussian noise terms. To ensure physical consistency between the generated data and the observed projections, a data consistency constraint is imposed by correcting the updated components in the projection domain using the projection operator $P(\cdot)$, as follows:
\begin{equation}
\bar{y}_{t-1}^{\omega} \leftarrow \bar{y}_{t-1}^{\omega} - P(\bar{y}_{t-1}^{\omega}) + P(y_{s}^{\omega}), \quad \omega \in \{L, H\},
\label{eq24}
\end{equation}
where $y_{s}^{\omega}$ denotes the observed low-frequency and high-frequency components, respectively. This multi-frequency decomposition, combined with independent iterative refinement, effectively aligns frequency-domain information and improves the reconstruction of fine details.

After correcting the projection data across multiple frequency bands, the inverse stationary wavelet transform (ISWT) is applied to fuse the low-frequency approximation component and the high-frequency detail components from different directions, reconstructing the corrected full projection data. This process is expressed as:
\begin{equation}
\bar{y}_{0} = \bar{y}^{L}_{0} + \sum_{i=1}^{3} \bar{y}_{0}^{H_{i}},
\label{eq:iSWT}
\end{equation}
where $\bar{y}^{L}_{0}$ denotes the corrected low-frequency approximation component, and $\bar{y}_{0}^{H_{}i}$ represent the high-frequency detail components along the $i$-th direction.

Subsequently, FBP is employed to reconstruct the final image from the corrected projection data. The projection data at each angle $\theta$ is first filtered via convolution:
\begin{equation}
q_{\theta}(r) = (h * \bar{y}_{0,\theta})(r),
\label{eq:filter}
\end{equation}
where $h$ is the filter function, and $r$ denotes the detector coordinate. The reconstructed image at spatial coordinate $(x,y)$ is then given by the backprojection integral:
\begin{equation}
\bar{x}(x,y) = \int_{\theta_{min}}^{\theta_{max}} \frac{1}{D^{2} + r^{2}} q_{\theta}(r)  d\theta,
\label{eq:filter}
\end{equation}
where the detector coordinate $r$ is determined by the following geometric relationship:
\begin{equation}
r = D \cdot \frac{x \cos \theta + y \sin \theta}{D - (x \sin \theta - y \cos \theta)},
\label{eq:geometric}
\end{equation}
where $D$ is the distance from the X-ray source to the rotation center. The weighting factor $\frac{1}{D^{2} + r^{2}}$ accounts for the diverging geometry of the fan-beam system.

Ultimately, the back-projection of the wavelet domain refined projection data yields a reconstructed image with superior structural coherence and high-fidelity detail preservation.

\begin{algorithm}[H]
\caption{Training and Iterative Reconstruction Process.}
\begin{algorithmic}[1]
\STATE \textbf{Input:} Training data $\mathcal{D}$, noise schedule $\{\alpha_t\}$, 
       guidance indicator $\delta \sim \text{Bernoulli}(p)$, masked guidance $\tilde{c}_M$, sparse-view projection $\hat{y}$;
\STATE \textbf{Output:} Trained models $\epsilon_\theta, s_{\theta_L}, s_{\theta_H}$ and reconstructed image $\bar{x}$.

\STATE \textbf{Training Stage}
\STATE Train $\epsilon_\theta, s_{\theta_L}, s_{\theta_H}$ on $\mathcal{D}$ using noisy projections and optional guidance $\tilde{c}_M$.

\STATE \textbf{Iterative Reconstruction Stage}
\FOR{$t=T-1$ down to $0$}
    \STATE Compute guidance strength, predict noise (Eq.\ref{eq17}, \ref{eq18});
    \STATE Update $y_{t-1}$ via DDIM reverse step (Eq.\ref{eq19});
\ENDFOR
\STATE compute $\{a^*,b^*\}$ and correct $\bar{y}_{T} \leftarrow a^*y_{0}+b^*$;

\STATE Decompose $\bar{y}$ into $\{\bar{y}^{L},\bar{y}^{H} \}$ via SWT;
\FOR{$t=T-1$ to $0$}
    \STATE Update low/high-frequency: $\bar{y}_t^L \leftarrow s_{\theta_L}(\bar{y}_{t-1}^L)$, $\bar{y}_t^H \leftarrow s_{\theta_H}(\bar{y}_{t-1}^H)$;
    \STATE Enforce data consistency (Eq.\ref{eq24});
\ENDFOR

\STATE Reconstruct $\bar{y}_{0} \leftarrow \text{ISWT}(\bar{y}_{0}^{L}, \bar{y}_{0}^{H})$;
\STATE Final reconstruction: $\bar{x} \leftarrow \mathrm{FBP}(\bar{y}_0)$;
\STATE \textbf{return} $\epsilon_\theta, s_{\theta_L}, s_{\theta_H}, \bar{x}$.
\end{algorithmic}
\label{alg:compact}
\end{algorithm}

\section{Experiment}
\subsection{Data Specification}
\subsubsection{Mayo 2016 Dataset}

We utilized the ``2016 NIH-AAPM Mayo Clinic Low-Dose CT Grand Challenge'' dataset \cite{chen2016open}   for training and testing, which contains 5,936 normal-dose CT slices (1 mm thickness) from 10 patients. Nine patients were used for training and one for testing.  Reference images were generated via filtered back-projection (FBP) with 720 projections. Fan-beam CT projections were simulated using a ray-driven algorithm. The rotational center-to-source and center-to-detector distances were both 40 cm. The detector width was 41.3 cm with 720 elements and evenly spaced 720 projections.

\subsubsection{Dental Arch Dataset}

The data set is the clinical data collected by JIROX Dental CBCT device produced by  YOFO (Hefei) Medical Technology Co., Ltd.  The source-to-image distance (SOD) of the device is 1700 mm, and the source-to-detector distance (SDD) is 1500 mm.  The device operates at a voltage of 100 kV and a current of 6 mA. The detector array consisted of 768×768 elements, with each detector element measuring 0.2×0.2 mm. The dataset consists of 20 cases, and each case provides 200 slices. The sinograms have a size of  1536×1600, which correspond to 1600 uniformly-sampled projections within the angular range of $[0^\circ, 360^\circ]$.  The dataset consists of full-view sinogram and their corresponding FBP reconstructions, with 512×512 image matrix. One patient case was randomly selected for CT reconstruction.

\subsubsection{Piglet Dataset}

To further evaluate the generalization capability of different methods on datasets from different centers, we also employed a pig CT dataset acquired using a GE Discovery CT750 HD scanner \cite{yi2018sharpness}. The dataset was scanned at 100 kVp and a slice thickness of 0.625 mm, with a dose of 300 mAs (conventional full dose), comprising a total of 850 images sized 512×512 pixels. A subset of slices from the dataset was used for testing.
  % Two slices were randomly selected from each anatomical region, resulting in 60 slices used for testing. 

\subsubsection{Mouse DECT Dataset}
 
This study used an actual raw projection dataset of anesthetized rats acquired in June 2025 by the Institute of Jinan Laboratory of Applied Nuclear Science. This scanning complied with the guidelines for ethical review of animal welfare and has obtained approval for animal experiments from the Ethics Committee of Nanchang University (Approvalnumber: 20220726008). The dataset was collected using a MARS multi-energy CT system, covering two energy ranges (21-30 kVp and 30-60 kVp), and comprised a total of 2,835 slices. All data were processed to match the training dataset. For experimental evaluation, a subset of slices was randomly selected from different anatomical regions. The scanning parameters were as follows: the source-to-object distance (SOD) was 76.28 mm, the source-to-detector distance (SDD) was 361.1 mm, and the detector pixel size was 0.1 mm.

\subsection{Implementation Details}  % 参数记得修改

In the experimental section of this study, the correction stage of the model was trained using the Adam optimizer with a learning rate of $10^{-4}$. The network was trained with 64 channels. Referring to previous studies, the conditional probability is set to  0.2.  A linear noise scheduling strategy was employed. The coarse generation stage was iterated 100 times, followed by 600 iterations in the correction stage. The method was implemented in Python using the Operator Discretization Library (ODL) \cite{rajpurkar2022ai} and the PyTorch framework. Experiments were conducted on a personal workstation equipped with a Tesla V100-PCIE 16GB GPU. Our source code is available on GitHub:  https://github.com/yqx7150/STRIDE.

Quantitative evaluation was performed using standard metrics (PSNR, SSIM, MSE).  The proposed algorithm was evaluated and compared with representative methods, including FBP \cite{brenner2007computed}, FBPConvNet \cite{jin2017deep}, RED-CNN \cite{yi2022red}, HDNet \cite{hu2020hybrid}, GMSD \cite{guan2023generative}, DPS \cite{chung2022diffusion}, GOUB \cite{yue2023image} and SWORD \cite{xu2024stage}. Sparse-view CT reconstruction was performed under varying projection numbers.

\begin{table*}[!t]
\caption{Reconstruction PSNR/SSIM/MSE ($10^{-3}$) for Mayo 2016 Dataset, Dental Arch Dataset, Piglet Dataset and Mouse DECT Dataset Using Different Methods at 60 Views.}
\centering
\label{view60_table}
\begin{tabularx}{\linewidth}{c|*{3}{>{\centering\arraybackslash}X}|
                                      *{3}{>{\centering\arraybackslash}X}|
                                      *{3}{>{\centering\arraybackslash}X}|
                                      *{3}{>{\centering\arraybackslash}X}}
\toprule
\multirow{2}{*}{\textbf{Method}} & \multicolumn{3}{c|}{\textbf{Mayo 2016 Dataset}} &  
\multicolumn{3}{c|}{\textbf{Dental Arch Dataset}} & 
\multicolumn{3}{c|}{\textbf{Piglet Dataset}} & 
\multicolumn{3}{c}{\textbf{Mouse DECT Dataset}} \\
\cmidrule(lr){2-4} \cmidrule(lr){5-7} \cmidrule(lr){8-10} \cmidrule(lr){11-13} 
 & PSNR($\uparrow$) & SSIM($\uparrow$) & MSE($\downarrow$) 
 & PSNR($\uparrow$) & SSIM($\uparrow$) & MSE($\downarrow$) 
 & PSNR($\uparrow$) & SSIM($\uparrow$) & MSE($\downarrow$) 
 & PSNR($\uparrow$) & SSIM($\uparrow$) & MSE($\downarrow$) \\
\midrule
FBP \cite{brenner2007computed} & 24.14 & 0.5626 & 3.889 & 25.65 & 0.6470 & 2.778 & 16.09 & 0.5419 & 3.889 & 24.83 & 0.6516 & 3.351 \\
FBPConvNet \cite{jin2017deep} & 35.57 & 0.9736 & 0.298 & 34.31 & 0.9707 & 0.436 & 32.71 & 0.9442 & 0.608 & 31.77 & 0.9351 & 0.681 \\
RED-CNN \cite{yi2022red} & 35.72 & 0.9290 & 0.317 & 34.10 & 0.9360 & 0.425 & 31.87 & 0.8756 & 0.723 & 29.25 & 0.7397 & 1.232 \\
HDNet \cite{hu2020hybrid} & 30.50 & 0.9008 & 1.746 & 29.09 & 0.9222 & 1.573 & 30.56 & 0.9015 & 1.685 & 28.41 & 0.8305 & 1.998 \\
GMSD \cite{guan2023generative} & 36.05 & 0.9642 & 0.267 & 34.91 & 0.9653 & 0.339 & 32.91 & 0.9404 & 0.553 & \underline{32.22} & \underline{0.9453} & \underline{0.657} \\
DPS \cite{chung2022diffusion} & 31.46 & 0.8137 & 0.832 & 32.72 & 0.8945 & 0.662 & 28.07 & 0.7519 & 1.749 & 28.60 & 0.8162 & 1.572 \\
GOUB \cite{yue2023image} & \underline{37.36} & 0.9439 & \underline{0.207} & 35.22 & 0.9527 & 0.344 & 32.06 & 0.8937 & 0.673 & 31.66 & 0.8375 & 0.716 \\
SWORD \cite{xu2024stage} & 36.80 & \underline{0.9734} & 0.223 & \underline{36.16} & \underline{0.9762} & \underline{0.259} & \underline{33.47} & \underline{0.9450} & \underline{0.509} & 31.69 & 0.9323 & 0.704 \\
\midrule
STRIDE (\textbf{ours}) & \textbf{39.14} & \textbf{0.9804} & \textbf{0.139} & \textbf{38.74} & \textbf{0.9870} & \textbf{0.144} & \textbf{35.67} & \textbf{0.9674} & \textbf{0.305} & \textbf{34.31} & \textbf{0.9597} & \textbf{0.421} \\
\bottomrule
\end{tabularx}
\end{table*}

\begin{figure*}[htbp]
    \centering
    % 在表格下方插入图片，宽度为\textwidth
    \includegraphics[width=1\textwidth]{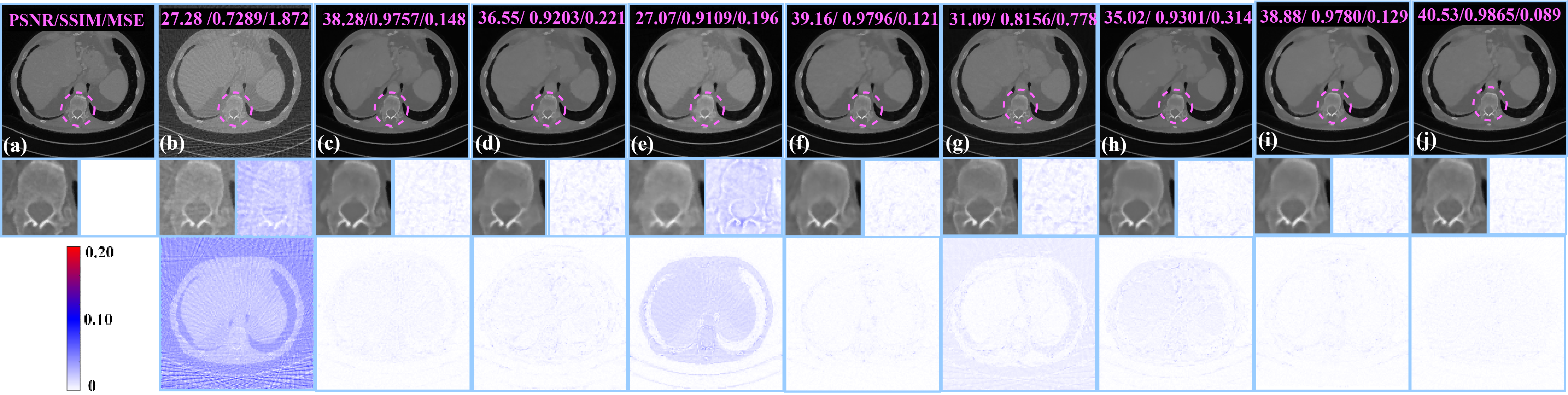} 
    \caption{ Reconstruction images from 60 views sparse CT using the Mayo 2016 dataset. (a) The reference image is compared with reconstructions obtained by (b) FBP, (c) FBPConvNet, (d) RED-CNN, (e) HDNet, (f) GMSD, (g) DPS, (h) GOUB, (i) SWORD, and (j) STRIDE. The second row shows magnified local regions to highlight the reconstruction of fine structural details. The third row presents the residual maps relative to the reference image, illustrating differences in artifact suppression, structural fidelity, and detail preservation across methods.}
    \label{MAYO_PIC}
\end{figure*}

\begin{table*}[h]
\centering
\caption{Reconstruction PSNR/SSIM/MSE ($10^{-3}$) for Mayo 2016 Dataset, Dental Arch Dataset, Piglet Dataset and Mouse DECT Dataset Using Different Methods at 72 Views.}
\label{view72_table}
\begin{tabularx}{\textwidth}{c|
    *{3}{>{\centering\arraybackslash}X}|
    *{3}{>{\centering\arraybackslash}X}|
    *{3}{>{\centering\arraybackslash}X}|
    *{3}{>{\centering\arraybackslash}X}}
\toprule
\multirow{2}{*}{\textbf{Method}} & \multicolumn{3}{c|}{\textbf{Mayo 2016 Dataset}} & 
\multicolumn{3}{c|}{\textbf{Dental Arch Dataset}} & 
\multicolumn{3}{c|}{\textbf{Piglet Dataset}} & 
\multicolumn{3}{c}{\textbf{Mouse DECT Dataset}} \\
\cmidrule(lr){2-4} \cmidrule(lr){5-7} \cmidrule(lr){8-10} \cmidrule(lr){11-13} 
 & PSNR($\uparrow$) & SSIM($\uparrow$) & MSE($\downarrow$) 
 & PSNR($\uparrow$) & SSIM($\uparrow$) & MSE($\downarrow$) 
 & PSNR($\uparrow$) & SSIM($\uparrow$) & MSE($\downarrow$) 
 & PSNR($\uparrow$) & SSIM($\uparrow$) & MSE($\downarrow$) \\
\midrule
FBP \cite{brenner2007computed} & 25.26 & 0.6147 & 3.001 & 26.77 & 0.6990 & 2.150 & 25.52 & 0.6219 & 2.933 & 25.54 & 0.7011 & 2.830 \\
FBPConvNet \cite{jin2017deep} & 37.77 & 0.9773 & 0.173 & 36.13 & 0.9749 & 0.266 & 35.32 & \underline{0.9647} & 0.334 & 33.02 & 0.9438 & 0.512 \\
RED-CNN \cite{yi2022red} & 36.32 & 0.9329 & 0.254 & 35.64 & 0.9508 & 0.306 & 34.15 & 0.9007 & 0.417 & 30.31 & 0.7934 & 0.961 \\
HDNet \cite{hu2020hybrid} & 31.66 & 0.9017 & 1.126 & 31.25 & 0.9212 & 0.958 & 31.97 & 0.8868 & 0.810 & 29.31 & 0.8293 & 1.597 \\
GMSD \cite{guan2023generative} & 38.08 & 0.9765 & 0.162 & 36.61 & 0.9752 & 0.231 & 35.11 & 0.9414 & 0.334 & \underline{34.14} & \underline{0.9546} & \underline{0.405} \\
DPS \cite{chung2022diffusion} & 34.12 & 0.8789 & 0.435 & 32.98 & 0.9125 & 0.592 & 30.99 & 0.8240 & 1.151 & 25.46 & 0.7351 & 3.058 \\
GOUB \cite{yue2023image} & 38.94 & 0.9480 & 0.130 & 36.26 & 0.9475 & 0.262 & 34.81 & 0.8721 & 0.375 & 32.48 & 0.8472 & 0.597 \\
SWORD \cite{xu2024stage} & \underline{39.55} & \underline{0.9835} & \underline{0.117} & \underline{39.40} & \underline{0.9870} & \underline{0.132} & \underline{35.90} & 0.9641 & \underline{0.296} & 32.92 & 0.9433 & 0.530 \\
\midrule
STRIDE (\textbf{ours}) & \textbf{40.37} & \textbf{0.9850} & \textbf{0.099} & \textbf{40.91} & \textbf{0.9910} & \textbf{0.086} & \textbf{37.89} & \textbf{0.9784} & \textbf{0.192} & \textbf{35.89} & \textbf{0.9675} & \textbf{0.288} \\
\bottomrule
\end{tabularx}
\end{table*}

\begin{figure*}[htbp]
    \centering
    % 在表格下方插入图片，宽度为\textwidth
    \includegraphics[width=1\textwidth]{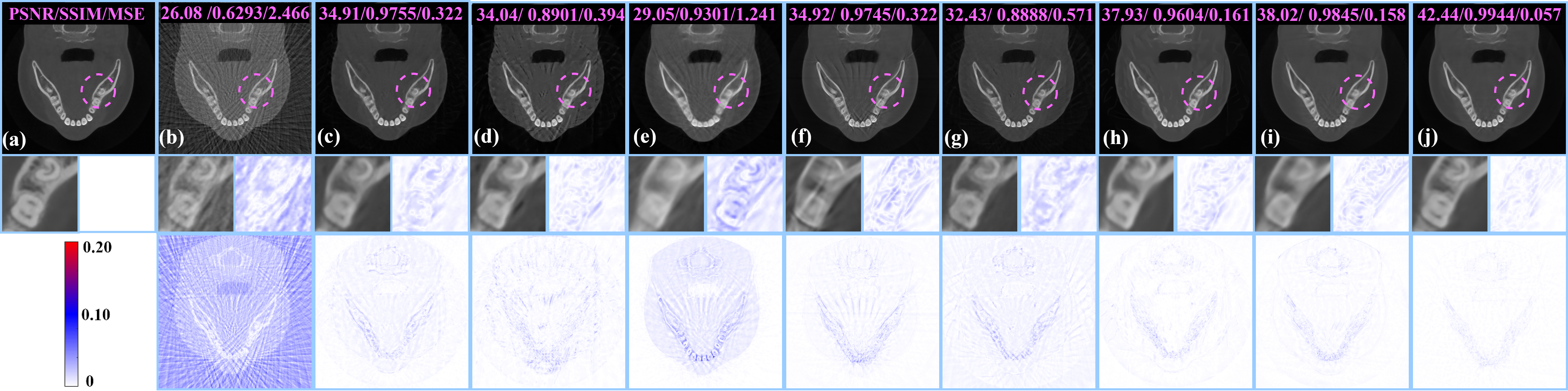} 
    \caption{Reconstruction images from 60 views sparse CT using the Dental Arch dataset. (a) The reference image is compared with reconstructions obtained by (b) FBP, (c) FBPConvNet, (d) RED-CNN, (e) HDNet, (f) GMSD, (g) DPS, (h) GOUB, (i) SWORD, and (j) STRIDE. The second row shows magnified local regions to highlight the reconstruction of fine structural details. The third row presents the residual maps relative to the reference image, illustrating differences in artifact suppression, structural fidelity, and detail preservation across methods.  }
    \label{DENTAL_PIC}
\end{figure*}

\begin{table*}[h]
\centering
\caption{Reconstruction PSNR/SSIM/MSE ($10^{-3}$) for Mayo 2016 Dataset, Dental Arch Dataset, Piglet Dataset and Mouse DECT Dataset Using Different Methods at 90 Views.}
\label{view90_table}
\begin{tabularx}{\textwidth}{c|
    *{3}{>{\centering\arraybackslash}X}|
    *{3}{>{\centering\arraybackslash}X}|
    *{3}{>{\centering\arraybackslash}X}|
    *{3}{>{\centering\arraybackslash}X}}
\toprule
\multirow{2}{*}{\textbf{Method}} & \multicolumn{3}{c|}{\textbf{Mayo 2016 Dataset}} & 
\multicolumn{3}{c|}{\textbf{Dental Arch Dataset}} & 
\multicolumn{3}{c|}{\textbf{Piglet Dataset}} & 
\multicolumn{3}{c}{\textbf{Mouse DECT Dataset}} \\
\cmidrule(lr){2-4} \cmidrule(lr){5-7} \cmidrule(lr){8-10} \cmidrule(lr){11-13} 
 & PSNR($\uparrow$) & SSIM($\uparrow$) & MSE($\downarrow$) 
 & PSNR($\uparrow$) & SSIM($\uparrow$) & MSE($\downarrow$) 
 & PSNR($\uparrow$) & SSIM($\uparrow$) & MSE($\downarrow$) 
 & PSNR($\uparrow$) & SSIM($\uparrow$) & MSE($\downarrow$) \\
\midrule
FBP \cite{brenner2007computed} & 27.14 & 0.6892 & 1.945 & 28.58 & 0.7629 & 1.414 & 27.14 & 0.6892 & 1.945 & 27.45 & 0.7574 & 1.834 \\
FBPConvNet \cite{jin2017deep} & 39.42 & 0.9847 & 0.121 & 37.92 & 0.9819 & 0.177 & 37.47 & 0.9736 & 0.212 & 34.34 & 0.9510 & 0.378 \\
RED-CNN \cite{yi2022red} & 38.83 & 0.9515 & 0.138 & 36.95 & 0.9454 & 0.216 & 36.34 & 0.9276 & 0.268 & 33.38 & 0.8484 & 0.475 \\
HDNet \cite{hu2020hybrid} & 33.25 & 0.9285 & 0.775 & 32.84 & 0.9444 & 0.655 & 32.75 & 0.9160 & 0.639 & 31.11 & 0.8626 & 1.041 \\
GMSD \cite{guan2023generative} & 39.64 & 0.9825 & 0.113 & 39.96 & 0.9896 & 0.106 & 37.49 & \underline{0.9850} & 0.193 & 35.01 & 0.9623 & 0.341 \\
DPS \cite{chung2022diffusion} & 34.94 & 0.8770 & 0.404 & 34.47 & 0.9209 & 0.496 & 31.03 & 0.8185 & 0.870 & 28.24 & 0.8045 & 6.314 \\
GOUB \cite{yue2023image} & 40.07 & 0.9578 & 0.103 & 38.34 & 0.9631 & 0.161 & 37.34 & 0.9440 & 0.212 & 34.19 & 0.8586 & 0.401 \\
SWORD \cite{xu2024stage} & \underline{41.55} & \underline{0.9885} & \underline{0.077} & \underline{42.76} & \underline{0.9940} & \underline{0.060} & \underline{39.18} & 0.9809 & \underline{0.145} & \underline{35.76} & \underline{0.9670} & \underline{0.288} \\
\midrule
STRIDE (\textbf{ours}) & \textbf{42.04} & \textbf{0.9888} & \textbf{0.069} & \textbf{43.33} & \textbf{0.9944} & \textbf{0.050} & \textbf{40.12} & \textbf{0.9851} & \textbf{0.097} & \textbf{37.44} & \textbf{0.9743} & \textbf{0.193} \\
\bottomrule
\end{tabularx}
\end{table*}

\begin{figure*}[htbp]
    \centering

    % 在表格下方插入图片，宽度为\textwidth
    \includegraphics[width=1\textwidth]{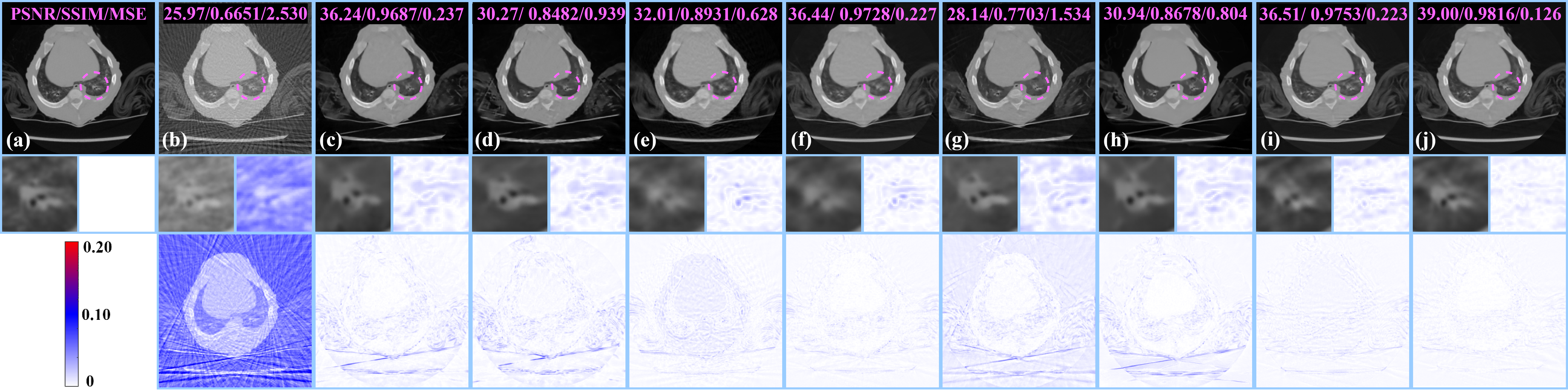} 

    \caption{Reconstruction images from 60 views sparse CT using the Piglet Dataset. (a) The reference image is compared with reconstructions obtained by (b) FBP, (c) FBPConvNet, (d) RED-CNN, (e) HDNet, (f) GMSD, (g) DPS, (h) GOUB, (i) SWORD, and (j) STRIDE. The second row shows magnified local regions to highlight the reconstruction of fine structural details. The third row presents the residual maps relative to the reference image, illustrating differences in artifact suppression, structural fidelity, and detail preservation across methods.}
    \label{PIGLET_PIC}
\end{figure*}

\begin{figure*}[htbp]
    \centering
    % 在表格下方插入图片，宽度为\textwidth
    \includegraphics[width=1\textwidth]{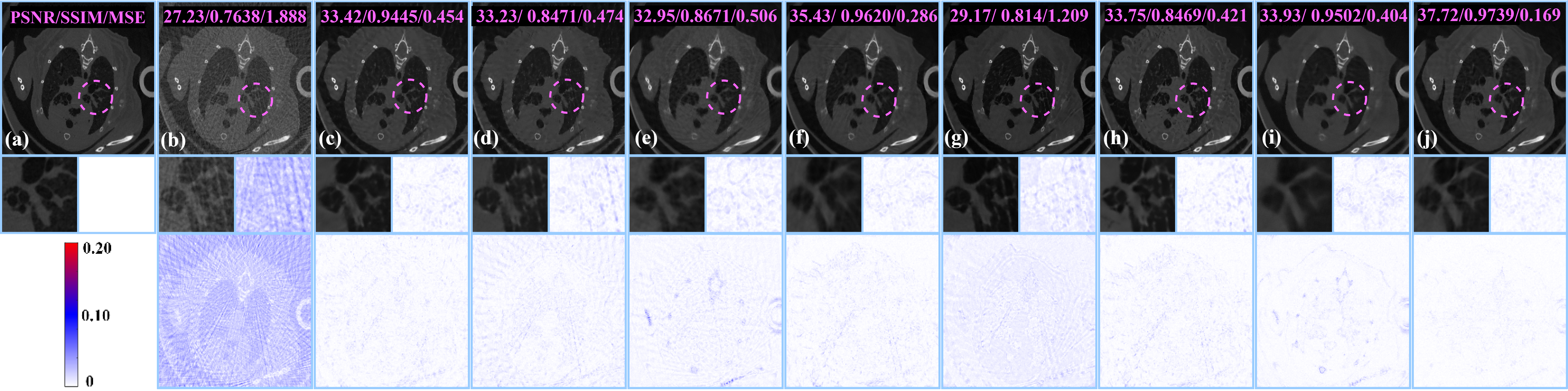} 
    \caption{Reconstruction images from 60 views sparse CT using the Mouse DECT Dataset. (a) The reference image is compared with reconstructions obtained by (b) FBP, (c) FBPConvNet, (d) RED-CNN, (e) HDNet, (f) GMSD, (g) DPS, (h) GOUB, (i) SWORD, and (j) STRIDE. The second row shows magnified local regions to highlight the reconstruction of fine structural details. The third row presents the residual maps relative to the reference image, illustrating differences in artifact suppression, structural fidelity, and detail preservation across methods.}
    \label{RAT_PIC}
\end{figure*}

\begin{table*}[!ht]
\centering
\caption{Quantitative evaluation PSNR/SSIM/MSE ($10^{-3}$) of network component ablations under 60, 72, and 90 views.}
\label{ABLATION_TABLE}

\begin{tabular}{@{}cccc|ccc|ccc|ccc@{}}
\toprule
\multirow{2}{*}{\textbf{GuidedNet}} & 
\multirow{2}{*}{\textbf{Consis}} & 
\multirow{2}{*}{\textbf{LowCorr}} & 
\multirow{2}{*}{\textbf{HighCorr}} & 
\multicolumn{3}{c|}{\textbf{60 Views}} & 
\multicolumn{3}{c|}{\textbf{72 Views}} & 
\multicolumn{3}{c}{\textbf{90 Views}} \\
\cmidrule(lr){5-7} \cmidrule(lr){8-10} \cmidrule(lr){11-13}  
 & & & & PSNR($\uparrow$) & SSIM($\uparrow$) & MSE($\downarrow$) 
 & PSNR($\uparrow$) & SSIM($\uparrow$) & MSE($\downarrow$) 
 & PSNR($\uparrow$) & SSIM($\uparrow$) & MSE($\downarrow$) \\
\midrule
\checkmark &  &  &  & 34.29 & 0.8975 & 0.410 & 35.60 & 0.9156 & 0.303  & 36.56  & 0.9312 & 0.241 \\  
\checkmark & \checkmark &  &  & 34.90 & 0.9145 & 0.358 & 35.94 & 0.9229 & 0.280  & 37.47  & 0.9388 & 0.193  \\  
\checkmark &  & \checkmark &  & 23.62 & 0.6026 & 4.748 & 23.98 & 0.6122 & 4.540  & 25.38  & 0.6805 & 3.204  \\ 
\checkmark &  &  & \checkmark & 12.87 & 0.3222 & 52.54 & 12.85 & 0.3322 & 52.81  & 12.82  & 0.2921 & 53.24  \\ 
\checkmark &  & \checkmark & \checkmark & 27.06 & 0.7719 & 1.986 & 27.90 & 0.7942 & 1.639 & 28.78 & 0.8238 & 1.337 \\  
\checkmark & \checkmark & \checkmark &  & 37.42 & 0.9730 & 0.194 & 39.81 & 0.9830 & 0.107 & 40.20 & 0.9841 & 0.102 \\
\checkmark & \checkmark &  & \checkmark & 36.98 & 0.9746 & 0.220 & 38.17 & 0.9804 & 0.171 & 39.76 & 0.9843 & 0.113 \\  
\checkmark & \checkmark & \checkmark & \checkmark & \textbf{39.14} & \textbf{0.9804} & \textbf{0.139} & \textbf{40.37} & \textbf{0.9850} & \textbf{0.099} & \textbf{42.04} & \textbf{0.9888} & \textbf{0.069}  \\   
\bottomrule
\end{tabular}
\end{table*}

\subsection{Reconstruction Experiments}
\subsubsection{Mayo 2016 Reconstruction Results} % 先表格再图像描述

Tables \ref{view60_table} to \ref{view90_table} present the quantitative evaluation results of various methods on the Mayo 2016 dataset. For reconstructed images with different projection views, the best values for each metric are highlighted in bold, while the second-best values are underlined. The STRIDE method outperforms all other compared methods across all metrics. Although methods such as FBPConvNet, GMSD, GOUB, and SWORD achieve relatively stable reconstruction results, they still show limitations in detail restoration. In contrast, STRIDE demonstrates significant advantages in improving image quality, effectively overcoming performance bottlenecks.

Fig. \ref{MAYO_PIC} presents the reconstructed visual results and corresponding error maps on the Mayo 2016 test dataset. FBP suffers from significant detail blurring and edge artifacts, compromising image quality. Although FBPConvNet improves image clarity, it still lacks edge sharpness and fails to fully restore fine structural details, resulting in the loss of subtle features. As a supervised image-domain learning method, RED-CNN exhibits over-smoothing during deblurring, which weakens detail recovery. The dual-domain iterative model HDNet enhances overall reconstruction quality but also causes excessive smoothing of structural details, leading to the loss of critical information. Similarly, DPS, GMSD, GOUB, and SWORD generally attenuate texture information during reconstruction, reducing realism and detail fidelity. In contrast, STRIDE excels at preserving texture details and restoring edge contours, showing the smallest residuals in the error maps and more accurately recovering key shape and grayscale features. Both qualitative and quantitative metrics demonstrate that STRIDE significantly outperforms other methods, highlighting its superiority in high-quality image reconstruction.

\subsubsection{Dental Arch Reconstruction Results}

To evaluate the practical applicability of the method, the model learns prior knowledge from the Mayo 2016 dataset and undergoes generalization validation on the Dental Arch dataset. Results in Tables \ref{view60_table} to \ref{view90_table} demonstrate that STRIDE significantly outperforms all comparative methods across quantitative metrics. In contrast, other methods consistently underperform relative to STRIDE, with the gap widening as the number of projection views decreases. STRIDE exhibits strong generalization and robustness, highlighting its superior performance.

As shown in Fig. \ref{DENTAL_PIC}, our method demonstrates superior visual reconstruction quality on the Dental Arch dataset. In contrast, FBP, RED-CNN, HDNet, GMSD, and GOUB exhibit prominent streak artifacts that cause loss of structural information, negatively impacting diagnostic accuracy. Although FBPConvNet, DPS, and SWORD partially mitigate these artifacts, their reconstructions still suffer from over-smoothed details and blurring, especially in the molar region, where boundary delineation is unclear, hindering identification of critical anatomical structures. By comparison, STRIDE not only effectively suppresses streak artifacts but also excels at preserving texture details, enabling precise restoration of complex anatomical features. Specifically, our method clearly distinguishes between molars and alveolar bone and provides more refined and accurate depiction of the pulp boundaries. Overall, STRIDE achieves outstanding reconstruction quality and structural fidelity in oral imaging, offering more reliable and precise support for clinical diagnosis and treatment planning.

\subsubsection{Piglet Reconstruction Results}

This study systematically evaluated the generalization performance of the proposed method on the piglet dataset. As shown in Tables \ref{view60_table} to \ref{view90_table}, STRIDE consistently outperformed comparative methods across all quantitative metrics, demonstrating superior generalization and robustness.  Fig. \ref{PIGLET_PIC} illustrates that FBPConvNet, RED-CNN, DPS, and GOUB exhibited over-smoothing during artifact suppression and image restoration, resulting in distortion and loss of critical anatomical details such as fine pulmonary structures. Although FBP, HDNet, GMSD, and SWORD preserved partial original structural information, their reconstructions suffered from blurred details and insufficient image quality, limiting accurate anatomical delineation. In contrast, STRIDE effectively suppressed streak artifacts and excelled at preserving texture details, significantly enhancing structural integrity and visual quality. It accurately restored complex pulmonary microstructures, maintaining high consistency with the ground truth.

\subsubsection{Mouse DECT Reconstruction Results}

This study evaluated the generalization performance on real scanned mouse data. Quantitative results in Tables \ref{view60_table} to \ref{view90_table} and Fig. \ref{RAT_PIC} indicate that  GMSD  showed notable performance improvements on the mouse dataset, ranking second best. In contrast, STRIDE consistently achieved the best numerical performance, further confirming its superiority and robustness. Visually, image-domain reconstruction methods such as FBPConvNet, RED-CNN, DPS, and GOUB produced prominent artifacts that deviated from true anatomical structures, compromising image fidelity. Other comparative methods retained partial structural information but exhibited local detail blurring, limiting accurate identification of key anatomical features. In comparison, STRIDE excelled in detail restoration and structural fidelity, significantly enhancing reconstruction accuracy and image quality.

%\begin{figure}[!t]
%\centering
%\includegraphics[width=3.5in]{Netablation.png}
%\caption{Ablation study of the STRIDE framework. (a) Ground truth. Reconstructions with individual modules: (b) temporally reweighted network, (c) adaptive linear data-consistency correction, (d) low-frequency network, (e) high-frequency network. Module combinations reveal synergy: (f) low+high-frequency correction, (g) without low-frequency, (h) without high-frequency. (i) Full STRIDE model achieves superior structural fidelity, artifact suppression, and fine-detail recovery, highlighting the unique contribution of each component.}
%\label{Net_PIC}
%\end{figure}

\subsection{Ablation  Study}

To systematically evaluate the practical efficacy of each component, we designed and conducted ablation experiments. By testing model components one by one, we precisely quantified the contribution of each module to the overall performance, ensuring the comprehensiveness and reliability of the evaluation results.

\subsubsection{Validation of Different Model Components}

To investigate the impact of different model component combinations on reconstruction performance, we designed a series of ablation experiments. To ensure fair comparison, all experiments used the same number of sampling steps as a unified baseline. Table \ref{ABLATION_TABLE} %and Fig. \ref{Net_PIC} 
present the effects of component stacking on reconstruction results. The results indicate significant functional differences among components: the guidance mechanism enables rapid preliminary adjustment of data distribution, the linear corrector precisely rectifies global distribution bias, the low-frequency corrector further refines offset adjustments, and the high-frequency corrector performs fine correction on difficult-to-recover high-frequency details. Overall, STRIDE achieves substantial performance improvement through the collaborative optimization of all components, outperforming all baseline methods and demonstrating excellent adaptability and robustness.

\subsubsection{Ablation Analysis Across Different Experimental Settings}

\begin{figure}[!t]
\centering
\includegraphics[width=3.5in]{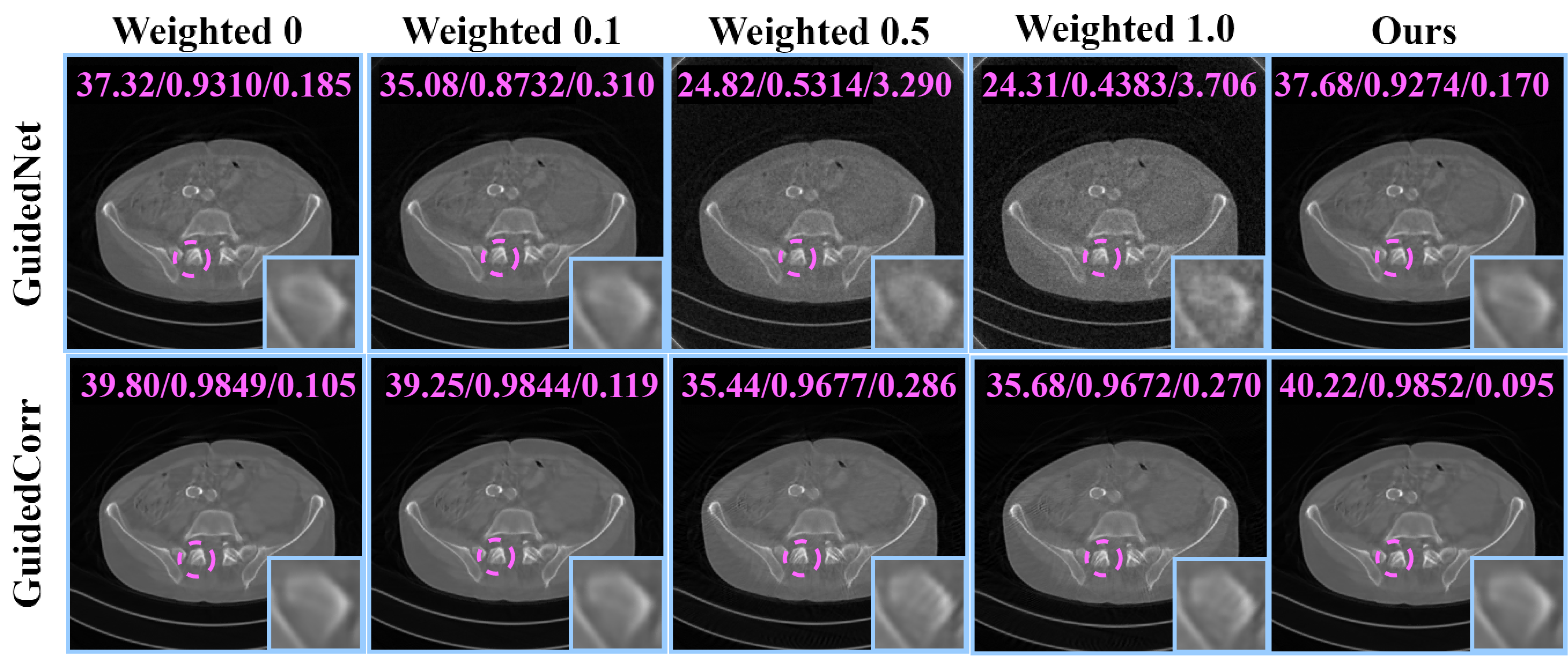}
\caption{Effect of different guidance weighting strategies. The first row compares reconstructions without weighting, with weights 0.1 and 0.5, and with the proposed temporally reweighted strategy. The second row shows the combined effect of the first stage with the second-stage refinement, demonstrating the efficacy of the temporally reweighted guidance in enhancing reconstruction fidelity.}
\label{weighted_PIC}
\end{figure}

\begin{figure}[!t]
\centering
\includegraphics[width=3.5in]{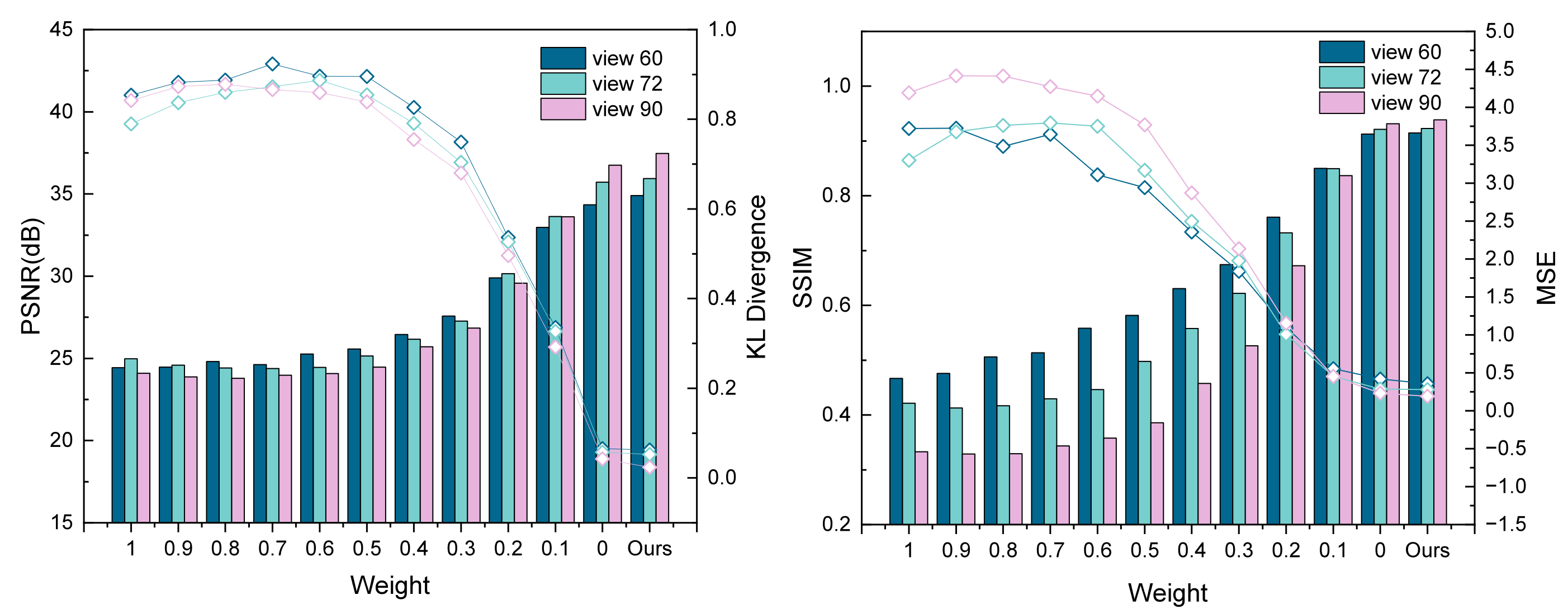}
\caption{Impact of guidance weighting on reconstruction and distribution alignment. PSNR, SSIM, MSE and KL divergence  are evaluated for weights from 0 to 1 with step 0.1, with the final point representing the proposed STRIDE method. The results highlight the superior performance and stability of STRIDE, illustrating how temporally reweighted guidance effectively balances reconstruction fidelity and distribution consistency.}
\label{lambda_PIC}
\end{figure}

We conducted ablation experiments to evaluate different guidance mechanisms, specifically comparing fixed-weight guidance and temporally reweighted dynamic guidance. As shown in Fig. \ref{weighted_PIC} and Fig. \ref{lambda_PIC}, increasing the fixed guidance weight impairs the network’s ability to learn data distribution features and noise patterns, resulting in degraded reconstruction performance. In contrast, the time-based dynamic guidance method follows the evolution of stochastic noise, effectively preserving the network’s noise prediction accuracy and progressively guiding the model to generate data distributions that are statistically consistent.

\section{Expansion of Downstream Tasks}

\begin{figure}[!t]
\centering
\includegraphics[width=3.5in]{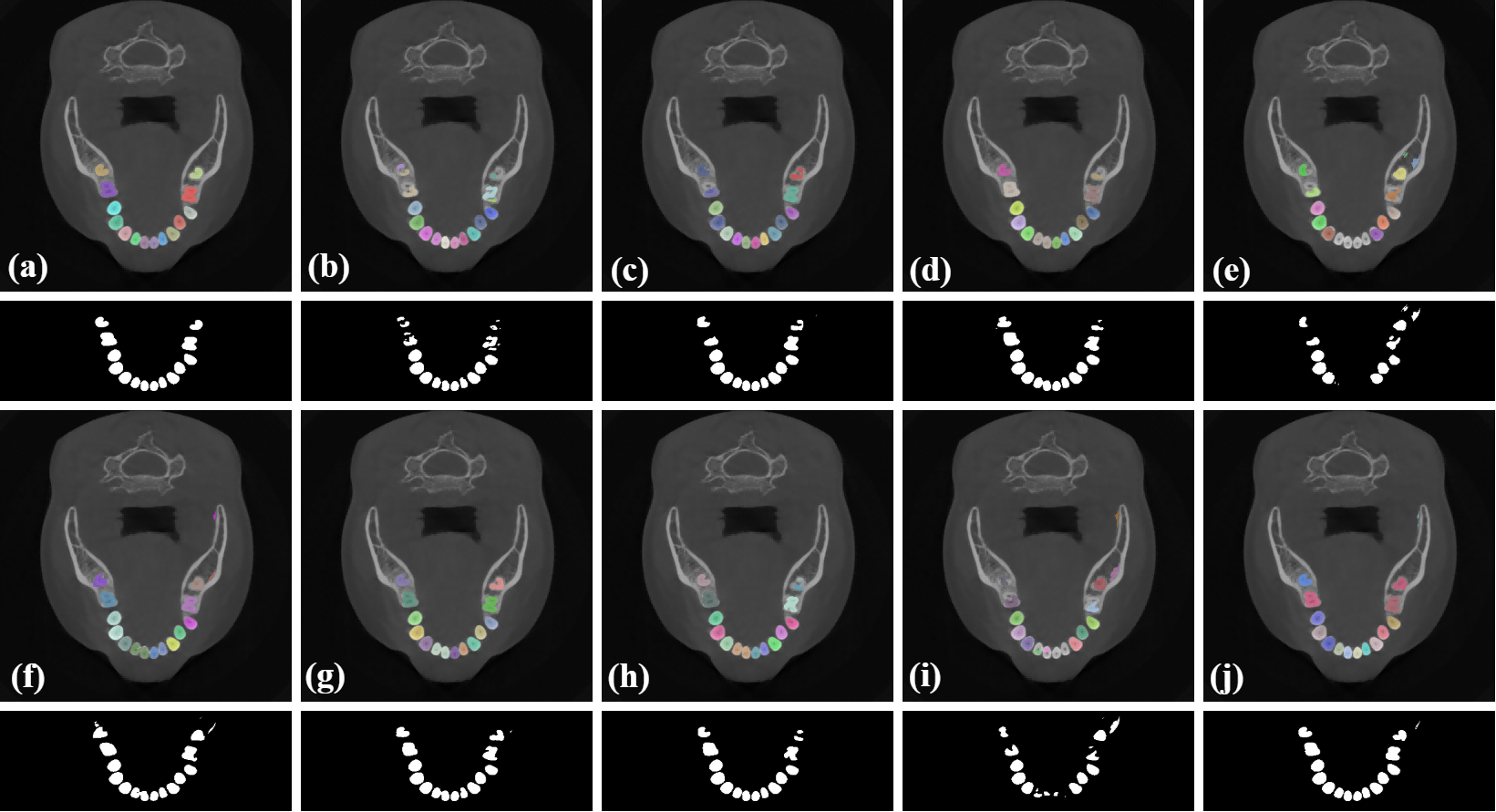}
\caption{Tooth segmentation results from 60 views sparse CT reconstructions using the Dental Arch dataset. (a) The reference segmentation obtained from the full-view reconstruction is compared with results obtained on 60-view reconstructions by (b) FBP, (c) FBPConvNet, (d) RED-CNN, (e) HDNet, (f) GMSD, (g) DPS, (h) GOUB, (i) SWORD, and (j) STRIDE. The second row shows the segmentation mask results obtained by each method.}
\label{tooth_seg_PIC}
\end{figure}

To evaluate the effectiveness of reconstructed images in downstream tasks, we conducted tooth segmentation experiments on the YOFO dataset using sparse-view reconstruction results. The evaluation metrics included DSC, IoU and Recall. The segmentation model was a U-Net \cite{ronneberger2015u} trained on the ToothFairy2 dataset \cite{2025CVPR,2024TMI, 2024IEEEACCESS}, with ground truth for the test set obtained from full-sampling segmentation results. As shown in Table \ref{tooth_seg_TABLE} and Fig. \ref{tooth_seg_PIC}, all comparison methods except GMSD suffered from severe under-segmentation, failing to preserve the complete tooth structure, while GMSD exhibited over-segmentation, misclassifying large portions of alveolar bone as teeth, thereby reducing segmentation accuracy. In contrast, STRIDE not only outperformed all baselines across quantitative metrics but also more accurately distinguished teeth from surrounding tissues, preserving both boundaries and overall morphology, and thus achieved the most stable and superior performance in the downstream segmentation task.

\begin{table}[!ht]
\centering
\caption{Quantitative evaluation of tooth segmentation on reconstructions.}
\label{tooth_seg_TABLE}
\renewcommand{\arraystretch}{1.1} % 行高
\setlength{\tabcolsep}{5pt} % 列间距

\begin{tabular}{@{}ccccc@{}} 
\toprule[0.8pt]
\textbf{Method} & \textbf{DSC}($\uparrow$) & \textbf{IoU}($\uparrow$) & \textbf{Recall}($\uparrow$)  \\
\midrule[0.6pt]
FBP \cite{brenner2007computed}      & 0.9143 & 0.8441 & 0.8823  \\
FBPConvNet \cite{jin2017deep}       & 0.9202 & 0.8554 & 0.8861  \\
RED-CNN \cite{yi2022red}             & \underline{0.9335} & \underline{0.8776} & \textbf{0.9282}  \\
HDNet \cite{hu2020hybrid}           & 0.6425 & 0.5076 & 0.5189  \\
GMSD \cite{guan2023generative}      & 0.8546 & 0.7539 & 0.7926  \\
DPS \cite{chung2022diffusion}       & 0.9157 & 0.8487 & 0.8871  \\
GOUB \cite{yue2023image}            & 0.9018 & 0.8265 & 0.8510  \\
SWORD \cite{xu2024stage}            & 0.8241 & 0.7118 & 0.7370  \\
\midrule[0.6pt]
STRIDE (\textbf{ours})              & \textbf{0.9433} & \textbf{0.8953} & \underline{0.9216}  \\
\bottomrule[0.8pt]
\end{tabular}
\end{table}

\section{Conclusion}

In this study, we propose a sparse conditional temporal reweighting distribution-estimation-guided diffusion model (STRIDE) for sparse-view CT reconstruction. At each iteration, an adjustment parameter $\lambda$ can be constructed based on the predicted result $\bar{x}*{0}$, effectively guiding the predictive distribution toward convergence with the original data $x*{0}$. Both theoretical derivations and empirical evaluations demonstrate that this strategy improves prediction accuracy and exhibits strong generalization capability.

By progressively optimizing the predicted results, STRIDE achieves a balance between fast coarse generation and refined correction. Experiments show that it substantially enhances global structural consistency and fine-detail reconstruction, improving the accuracy and diagnostic reliability of the reconstructed images. Evaluations across multiple datasets further confirm its superior generalization ability and robustness. Nevertheless, this study mainly focuses on empirical validation, and the design of $\lambda$ still requires task-specific optimization to fully exploit its potential.

%
%\section*{Acknowledgments}
%This work was supported by the Imaging and Visual Representation Laboratory, School of Information Engineering, Nanchang University. The authors would also like to express their sincere gratitude to YOFO Medical Technology Co., Ltd. and the Jinan Branch of the Institute of Jinan Laboratory of Applied Nuclear Science, for their support of this study.

\bibliographystyle{IEEEtran}
\bibliography{IEEEabrv,reference}

\clearpage

{\appendix}

\begin{lemma}[Cauchy-Schwarz Inequality]\label{lemma}
For any vectors $a, b \in \mathbb{R}^n$, the following inequality holds:
$ |\langle a, b \rangle| \leq \|a\| \cdot \|b\|$, with equality if and only if $a$ and $b$ are linearly dependent.
\end{lemma}

\section*{Proof of Theorem 1}

Let the initial signal estimate at timestep $t$ is given by:
\begin{equation}
\hat{y}_{0}^{(t)} = \frac{1}{\sqrt{\alpha}_{t}} (x_{t}-\sqrt{1-\alpha_{t}} \cdot \epsilon_{\theta}(y_{t},t)).  \label{eq_app_1}
\end{equation}
where $\alpha_{t}$ is the cumulative noise coefficient determined by the noise schedule, and $\epsilon_{\theta}$ denotes the model predicted additive noise component. Based on this, the corrected estimate can be expressed as:
\begin{equation}
\tilde{y}_{0}^{(t)}=\hat{y}_{0}^{(t)}+\lambda_{t}M(\hat{y}_{0}^{(t)}-y_{s}),  \label{eq_app_2}
\end{equation}
where $\lambda_{t}\in [0,1]$ is a time-dependent correction weight, and $y_{s}$ represents the observed data. 

Let the ground truth be denoted by $y_{g}$, and define the error terms as  $\zeta_{0}^{(t)} = \hat{y}_{0}^{(t)}-y_{g} $, $\xi_{g}=y_{s}-y_{g} $.  Then, the corrected error can be expressed as
\begin{equation}
\tilde{\delta}^{(t)} =\tilde{y}_{0}^{(t)}-y_{g}=(1-\lambda_{t}M)\zeta_{0}^{(t)} + \lambda_{t}M \xi_{g}.  \label{eq_app_4}
\end{equation}
We now derive the squared norm of the corrected error:
\begin{align}
 \| \tilde{\delta}^{(t)} \|^{2} &= \| (1-\lambda_{t}M)\zeta_{0}^{t} +\lambda_{t}M \xi_{g} \|^{2} \\
    &= (1 - \lambda_{t}M)^{2}\| \zeta_{0}^{(t)} \|^{2} + \lambda_{t}M^{2}\| \xi_{g} \|^{2}  \notag  \\ 
    &+2\lambda_{t}M(1-\lambda_{t}M) \langle \zeta_{0}^{(t)}, \xi_{g} \rangle. 
\label{eq_app_5}
\end{align}
By invoking Lemma~\ref{lemma}, we have $ | \langle \zeta_{0}^{(t)}, \xi_{g} \rangle | \leq \| \zeta_{0}^{(t)} \| \cdot \| \xi_{g} \|$, then  we can obtain the following upper bound:
\begin{align}
\| \tilde{\delta}^{(t)} \|^{2} & \leq (1-\lambda_{t})^{2}\|\zeta_{0}^{(t)} \|^{2}+ \lambda_{t}^{2}\| \xi_{g} \|^{2}  \notag   \\ 
    &+ 2\lambda_{t}(1-\lambda_{t})\|\zeta_{0}^{(t)}\| \cdot \| \xi_{g}\|.  
\label{eq_app_7}
\end{align}
Let  $a = \|\zeta_{0}^{(t)} \|$, $ b= \| \xi_{g} \| $,  expanding the terms yields:
\begin{equation}
\| \tilde{\delta}^{(t)} \|^{2} \leq (a^{2}+b^{2}-2ab)\lambda_{t}^{2} + (2ab-2a^{2}_{t})\lambda_{t} + a_{t}^{2}.  \label{eq_app_8}
\end{equation}
Let $f(\lambda)$ be defined as:
\begin{equation}
f(\lambda)=(a^{2}+b^{2}-2ab)\lambda^{2}+ (2ab-2a^{2}_{t})\lambda_{t}+a_{t}^{2}.  \label{eq_app_9}
\end{equation}
Since $a^{2}+b^{2}-2ab=(a-b)^{2}>0$, $f(\lambda)$ is convex with respect to $\lambda$,  its unique minimizer is:
\begin{equation}
 \lambda_{t}^{\ast}=\frac{a}{a-b}.  \label{eq_app_10}
\end{equation}
Consider the constrained minimization of $f(\lambda)$ over the interval $\lambda \in [0,1]$. 

If $\lambda_{t}^{\ast} \in [0,1]$ is the global minimizer of the objective function $f(\lambda)$ over this interval, then $f(\lambda_{t}^{\ast}) = \min_{\lambda \in [0,1]} f(\lambda)$. Consequently, for any fixed $\hat{\lambda}_{t} \in [0,1]$, it holds that 
$f(\lambda_{t}^{\ast}) \le f(\hat{\lambda}_{t})$. For all iterations, using the analytically computed optimal updates $\lambda_{t}^{\ast}$ at each step ensures that the accumulated objective value does not exceed that obtained using any fixed or manually chosen sequence of $\lambda_{t}$. Specifically, summing over all iterations yields:
\begin{equation}
 \sum_{t} f_{t}(\lambda_{t}^{\ast}) \leq \sum_{t}f_{t}(\hat{\lambda_{t}}), \  \forall \hat{\lambda_{t}} \in \mathbb{R}.  \label{eq_app_11}
\end{equation}
Therefore, when the closed-form solution $\lambda_{t}^{\ast}$ exists and satisfies the feasible constraint, it can be employed in place of manually tuned update schedules, ensuring a step-wise reduction of the objective function. \hfill $\square$

\section*{Proof of Corollary 1}

According to Theorem~\ref{thm1}, we are interested in determining the optimal correction weight $\lambda_t^\ast$ at each iteration step $t$. Consider the squared norm of the corrected error
$\|\tilde{\delta}^{(t)}\|^2 = \|(1-\lambda_t)\zeta_0^{(t)} + \lambda_t \xi_g\|^2$. Minimizing this error with respect to $\lambda_t$ leads to the optimal weight. Let $a_t^2 = \|\zeta_0^{(t)}\|^2$, $b^2 = \|\xi_g\|^2$, and $c_t = \langle \zeta_0^{(t)}, \xi_g \rangle$, we have $ \|\tilde{\delta}^{(t)}\|^{2} = (1-\lambda_t)^{2} a_{t}^{2} + \lambda_{t}^{2} b^{2} + 2 \lambda_{t} (1-\lambda_{t}) c_{t}$. Specifically, by differentiating $\|\tilde{\delta}^{(t)}\|^2$ with respect to $\lambda_t$,  While $\frac{d}{d\lambda_t} \|\tilde{\delta}^{(t)}\|^2 = -2 (1-\lambda_t) a_t^2 + 2 \lambda_t b^2 + 2 (1 - 2 \lambda_t) c_t = 0$, we can obtain the closed-form solution for the optimal correction weight, which defines the correction dynamics function:
\begin{equation}
\mathcal{F}(\lambda_t^\ast) = \frac{a_t^2 - c_t}{a_t^2 + b^2 - 2 c_t}. \label{eq_app_12}
\end{equation}
According to the Lemma~\ref{lemma}, we have $|c_t| = \left| \langle \zeta_{0}^{(t)}, \xi_g \rangle \right| \leq \left\| \zeta_{0}^{(t)} \right\| \cdot \left\| \xi_g \right\| = a_{t} b \leq b^2$ with strict inequality when $a_{t}<b$ .  Then we can obtain a bound on the inner product term $c_t$. To  analyze the monotonicity of $\lambda_{t}^{\ast}$  with respect to the model error magnitude $a_t^2$, compute the derivative of $\lambda_t^\ast$ as:
\begin{equation}
\frac{\partial \lambda_t^*}{\partial a_t^2} = \frac{b^2 - c_t}{(a_t^2 + b^2 - 2c_t)^2}.  \label{eq_app_13}
\end{equation}
Since the denominator $(a_t^2 + b^2 - 2 c_t)^2$ is strictly positive, the sign of the derivative is determined entirely by the numerator $b^2 - c_t$. 
\begin{itemize}
\item If $c_{t} < b^{2}$, then the derivative is positive, i.e., $\lambda_{t}^{\ast}$increases with  $a_{t}^{2}$ . Given that  $ a_t^2 = \|\zeta_0^{(t)}\|^2 $ decreases over time (due to iterative refinement), it follows that  $\lambda_{t}^{\ast}$  must also decrease with time $a_t^2 \geq a_{t+1}^2 \implies \lambda_{t}^{\ast} \geq \lambda_{t+1}^{\ast}$.  This proves monotonicity under the sufficient condition $\|\zeta_{0}\|^2 < \|\xi_{g}\|^2$.

\item If $c_{t} \ge b^{2}$, then $\frac{\partial \lambda_{t}^{\ast}}{\partial a_{t}^{2}} \le 0$.   According to Lemma~\ref{lemma}, we have $c_t \le a_t b \le b^2$, which contradicts the assumption that $c_t > b^2$.  For $c_{t} = b^{2}$, it indicates that the prediction error $\zeta_0^{(t)}$ is perfectly aligned with the observation error $\xi_g$,  the unconstrained optimal correction weight $\lambda_t^\ast$ may reach its upper bound. To ensure numerical stability and prevent over-correction, a truncation or constraint should be applied.
\end{itemize}
The monotonic decay of the correction coefficient $\lambda_{t}^{\ast}$over time is guaranteed if and only if $ \|\zeta_0\|^2 < \|\xi_g\|^2 $. When this condition is violated, the correction may act in the wrong direction, magnifying error and destabilizing the trajectory. Therefore, to ensure numerical stability and prevent divergence, a truncation or constraint on $\lambda_t^\ast$  should be applied. \hfill $\square$

%%%%%%%%%%%%%%%%%%%%%%%%%%%%%%%%%%%%%%%%%%%%%%%%%%%%%%%%%%%%%%%%%%%%%%%%%%%%%%%%%%%%%%%%%%%%%%

%\newpage
% your own custom macro containing the $\backslash${\tt{includegraphics}} command to make things
%\bf{If you include a photo:}\vspace{-33pt}
%\begin{IEEEbiography}[{\includegraphics[width=1in,height=1.25in,clip,keepaspectratio]{fig1}}]{Michael Shell}
%Use $\backslash${\tt{begin\{IEEEbiography\}}} and then for the 1st argument use $\backslash${\tt{includegraphics}} to declare and link the author photo.
%Use the author name as the 3rd argument followed by the biography text.
%\end{IEEEbiography}

%\vfill

\end{document}